\documentclass[accept,hidelinks,onefignum,onetabnum]{siamart250211}

\usepackage{etoolbox}
\makeatletter
\patchcmd{\@maketitle}{\uppercase}{\relax}{}{}
\patchcmd{\@ucnt}{\uppercase}{\relax}{}{}
\makeatother

\usepackage{lipsum}
\usepackage{amsfonts}
\usepackage{epstopdf}
\ifpdf
\DeclareGraphicsExtensions{.eps,.pdf,.png,.jpg}
\else
  \DeclareGraphicsExtensions{.eps}
\fi

\usepackage{microtype}
\usepackage{subfigure}
\usepackage{booktabs} 

\usepackage[textsize=tiny]{todonotes}

\usepackage{amsmath,amssymb,amsfonts,bm}
\usepackage{algorithm}
\usepackage{algorithmic}
\usepackage{graphicx}

\usepackage{hyperref}

\usepackage{textcomp}
\usepackage[T1]{fontenc}
\usepackage{color}
\usepackage{xurl}
\usepackage{bm}
\usepackage[normalem]{ulem}
\usepackage{xcolor}
\usepackage{xspace}
\usepackage{wrapfig}
\usepackage{booktabs, multirow}

\usepackage{enumitem}
\usepackage{Definitions}

\newtheorem{remark}{Remark}
\newtheorem{assumption}{Assumption}

\title{TS-RSR: a provably efficient approach for batch Bayesian Optimization
}

\author{Zhaolin Ren, Na Li\thanks{Zhaolin Ren and Na Li are with Harvard University, Email: {\tt\footnotesize zhaolinren@g.harvard.edu, nali@seas.harvard.edu}. This work is supported by NIH R01LM014465 and NSF AI institute: 2112085.
}
}

\begin{document}

\maketitle

\begin{abstract}
This paper presents a new approach for batch Bayesian Optimization (BO) called Thompson Sampling-Regret to Sigma Ratio directed sampling (TS-RSR),
where we sample a new batch of actions by minimizing a Thompson Sampling approximation of a regret to uncertainty ratio.
Our sampling objective is to
coordinate the actions chosen in each batch in a
way that minimizes redundancy between points whilst focusing on points with high predictive means or high uncertainty. Theoretically, we provide rigorous convergence guarantees on our algorithm's regret, and numerically, we demonstrate that our method attains state-of-the-art performance on a range of challenging synthetic and realistic test functions, where it outperforms several competitive benchmark batch BO algorithms.
    
\end{abstract}



\section{Introduction}\label{sec:intro}

Consider the following problem of batch Bayesian Optimization (batch BO). Let $\Xcal \subset \mathbb{R}^d$ be a bounded compact set. Suppose we wish to maximize an unknown function $f: \Xcal \to \mathbb{R}$, and our only access to $f$ is through a noisy evaluation oracle, i.e. $y = f(x) + \epsilon$, $\epsilon \sim N(0,\sigma_n^2)$, with $\sigma_n > 0$. We consider the batch setting, where we assume that we are able to query $f$ over $T$ rounds, and at each round, we can send out $m$ queries in parallel. We are typically interested in the case when $m > 1$, where we expect to do better than when $m = 1$. In particular, we are interested in quantifying the ``improvement'' that a larger $m$ can give us. 

To be more precise, let us discuss our evaluation metrics. Let $x_{t,i}$ denote the query point of the $i$-th agent at the $t$-th time. Let $x^* \in X$ denote a maximizer of $f$. In this paper, we provide bounds for the expected cumulative regret $\bbE[R_{T,m}]$, where
\begin{align*}
    R_{T,m} := \sum_{t=1}^T \sum_{i=1}^m [f(x^*) - f(x_{t,i})].
\end{align*}
We also define the simple regret as 
\begin{align*}
    S_{T,m} :=  \min_{t \in [T]}\min_{i \in [m]} f(x^*) - f(x_{t,i}),
\end{align*}
where we use the notation $[N] := \{1,2,\dots,N\}$ (for any positive integer $N$), which we will use throughout the paper. Note that the simple regret satisfies 
the relationship $S_{T,m} \leq \frac{1}{Tm} R_{T,m}$. This shows that a bound on the cumulative regret translates to a bound on the simple regret.

Without any assumptions on the smoothness and regularity of $f$, it may be impossible to optimize it in a limited number of samples; consider for instance functions that wildly oscillate or are discontinuous at many points. Thus, in order to make the problem tractable, we make the following assumption on $f$.

\begin{assumption}
    \label{assumption:GP}[GP model] The function $f$ is assumed to be a sample from a Gaussian Process (GP), where $\GP(0, k(\cdot,\cdot))$ is our GP prior over $f$. A Gaussian Process $\GP(\mu(x), k(x,x'))$ is specified by its mean function $\mu(x)= \xpt{f(x)}$ and covariance function $k(x,x') = \xpt{(f(x) - \mu(x))(f(x') - \mu(x'))}$. More details about Gaussian Processes can be found in \cite{williams2006gaussian}.
\end{assumption}


There are several existing algorithms for batch Bayesian optimization with regret guarantees, e.g. batch-Upper Confidence Bound (UCB) \cite{srinivas2009gaussian}, batch-Thompson sampling (TS) \cite{kandasamy2018parallelised}. There are known guarantees on the cumulative regret of batch-UCB and batch TS. Unfortunately, empirical performance of batch-UCB and batch-TS tend to be suboptimal. A suite of heuristic methods have been developed for batch BO, e.g. \cite{ma2023gaussian, garcia2019fully, gong2019quantile}. However, theoretical guarantees are typically lacking for these algorithms. This inspires us to ask the following question:
\begin{center}
    \textbf{Can we design theoretically grounded, effective batch BO algorithms \\that also satisfy rigorous guarantees?}
\end{center}

Inspired by the literature of information-directed sampling (IDS)  \cite{russo2014learning, baek2023ts}, we introduce a new algorithm for Bayesian Optimization (BO), which we call \emph{Thompson Sampling-Regret to Sigma Ratio directed sampling} ($\alg$). The algorithm works for any setting of the batch size $m$, and is thus also appropriate for batch BO. Our contributions are as follows.

First, on the algorithmic front, we propose a novel sampling objective for BO that automatically balances exploitation and exploration in a parameter-free manner (unlike for instance in UCB-type methods, where setting the confidence interval typically requires the careful, often domain-specific choice of a suitable hyperparameter). In particular, for batch BO, our algorithm is able to coordinate the actions chosen in each batch in an intelligent way that minimizes redundancy between points. Compared to standard batch UCB-type methods, the objective avoids the tuning of a confidence set hyperparameter, and compared to Thompson Sampling methods, we have improved redundancy avoidance.

Second, on the theoretical front, we show that under mild assumptions, the Bayesian cumulative regret $R_{T,m}$ of our algorithm (with an appropriate initialization strategy) scales as $\tilde{O}(\sqrt{\gamma_{Tm} Tm})$ where $T$ denotes the number of rounds and $m$ denotes the batch size, and $\gamma_{Tm}$ denotes a problem-dependent information-gain quantity. This matches with the best achievable rate with the same total number of function evaluations attained by standard sequential BO\footnote{In sequential BO, the function evaluation result is known after each evaluation, and there is no notion of batch}~\cite{srinivas2009gaussian}. 

Finally, empirically, we show via extensive experiments on a range of synthetic and real-world nonconvex test functions that our algorithm attains state-of-the-art performance in practice, outperforming other benchmark algorithms for batch BO. We have also published a version of our code at the following link:~\url{https://github.com/rafflesintown/TSRSR}.

\section{Related work}\label{sec:related}

There is a vast literature on Bayesian Optimization (BO) \cite{frazier2018tutorial} and batch BO. One popular class of methods for BO and batch BO is UCB-inspired methods, \cite{srinivas2009gaussian, desautels2014parallelizing,kaufmann2012bayesian, daxberger2017distributed}. Building on the seminal work in \cite{srinivas2009gaussian} which studied the use of the UCB acquisition function in BO with Gaussian Process and provided regret bounds, subsequent works have extended this to the batch setting. The most prominent approach in this direction is Batch UCB (BUCB) \cite{desautels2014parallelizing}, which is a sequential sampling strategy that keeps the posterior mean constant throughout the batch but updates the covariance as the batch progresses. Another notable work combines UCB with pure exploration by picking the first action in a batch using UCB and subsequent actions in the batch by maximizing posterior uncertainty. One key drawback of UCB-type methods is the strong dependence of empirical performance on the choice of the $\beta_t$ parameter; note that in UCB-type methods, the UCB-maximizing action is typically determined as $x_t^{\mathrm{UCB}} \in \argmax \mu_t(x) + \beta_t \sigma_t(x)$,  where $\beta_t$ shapes the weight allocation between the posterior mean $\mu_t$ and posterior uncertainty $\sigma_t$.  While there exist theoretically valid choices of $\beta_t$ that ensure convergence, practical implementations typically requiring heuristic tuning of the $\beta_t$ parameter. In contrast, in our algorithm, we do not require the tuning of such a $\beta_t$ parameter.

Another popular class of methods is Thompson Sampling (TS)-based methods \cite{kandasamy2018parallelised, dai2020federated,hernandez2017parallel}. The downside of TS-based methods is the lack of penalization for duplicating actions in a batch, which can result in over-exploitation and a lack of diversity, as discussed for instance in \cite{adachi_sober_2023}. On the other hand, as we will see, our method does penalize duplicating samples, allowing for better diversity across samples. 

In many practical applications, Expected Improvement (EI)~\cite{letham2019constrained,zhan2020expected,ament2024unexpected} is also a popular approach to BO. While EI lacks theoretical guarantees, it is popular amongst practioners of BO, and has also been extended to the batch setting, where it is also known as qEI~\cite{hunt2020batch,ginsbourger2008multi}.

In addition, informational approaches based on maximizing informational metrics have also been proposed for BO \cite{hennig2012entropy,hernandez2015predictive,wang2016optimization,wang2017max} and batch BO~\cite{shah2015parallel, garrido2019predictive,takeno2020multi, hvarfner2022joint}. While such methods can be effective for BO, efficient extension of these methods to the batch BO setting is a challenging problem, since the computational complexity of searching for a batch of actions that maximize information about (for instance) the location of the maximizer scales exponentially with the size of the batch. One interesting remedy to this computational challenge is found in \cite{ma2023gaussian}, which proposes an efficient gradient-descent based method that uses a heuristic approximation of the posterior maximum value by the Gaussian distribution for the output of the current posterior UCB. However, this method also relies on the tuning of the $\beta_t$ parameter in determining the UCB, and also does not satisfy any theoretical guarantees.

There are also a number of other works in batch BO which do not fall neatly into the categories above. These include an early work that tackles batch BO by trying to using Monte-Carlo simulation to select input batches that closely match the expected behavior of sequential policies \cite{azimi_batch_nodate}. However, being a largely heuristic algorithm, no theoretical guarantees exist. Other heuristic algorithms include an algorithm~\cite{gonzalez_batch_nodate} that proposes a batch sampling strategy that utilizes an estimate of the function's Lipschitz constant, Acquisition Thompson Sampling (ATS) \cite{de_palma_sampling_2019}, which is based on the idea of sampling multiple acquisition functions from a stochastic process, as well as an algorithm that samples according to the Boltzman distribution with the energy function given by a chosen acquisition function \cite{garcia-barcos_fully_2019}. However, being heuristics, these algorithms are not known to satisfy any rigorous guarantees. An interesting recent work proposes inducing batch diversity in batch BO by leveraging the Determinental Point Process (DPP) \cite{nava_diversified_2022}, and provides theoretical guarantees for their algorithm. However, a limitation of the algorithm is that the computational complexity of sampling scales exponentially with the number of agents, limiting the application of the algorithm for large batch problems. For large batch problems, there has been a very recent work~\cite{adachi_sober_2023} that seeks scalable and diversified batch BO by reformulating batch selection for global optimization as a quadrature problem. Nonetheless, this algorithm lacks theoretical guarantees, and being designed for large batch problems, e.g. $m$ in the hundreds, it may fail to be effective for moderate $m$ problems, e.g. $m$ less than $50$. Another interesting direction in batch BO considers the case when the delay in receiving the feedback of the function evaluation is stochastic~\cite{verma2022bayesian}; while orthogonal to our work, it could be meaningful to apply the methods proposed here to the stochastic delay batch BO setting.

Finally, we note that a strong inspiration on our work comes from ideas in the information directed sampling literature (e.g. \cite{russo2014learning, baek2023ts,kirschner2018information}), where the sampling at each stage also takes place based on the optimization of some regret to uncertainty ratio. While \cite{russo2014learning} and \cite{baek2023ts} did not cover the setting of BO with Gaussian Process (GP), we note that the algorithm in \cite{kirschner2018information} does apply to BO with GP, and they also provided high-probability regret bounds. However, the design of the sampling function in \cite{kirschner2018information} also requires choosing a $\beta_t$ parameter (similar to UCB type methods), which as we observed can be hard to tune in practice. 
\section{Problem Setup and Preliminaries}\label{sec:prelim}
Let $\Xcal \subset \mathbb{R}^d$ be a bounded compact set. Suppose we wish to maximize an unknown function $f: \Xcal \to \mathbb{R}$, and our only access to $f$ is through a noisy evaluation oracle, i.e. $y = f(x) + \epsilon$, $\epsilon \sim N(0,\sigma_n^2)$, with $\sigma_n > 0$. We assume that $f$ is drawn from a Gaussian process, as stated in Assumption~\ref{assumption:GP}. Let $x^* \in \mathcal{X}$ denote a maximizer of $f$. We consider the batch setting, where we assume that we are able to query $f$ over $T$ rounds, and at each round, we can send out $m$ queries in parallel. 

To streamline our analysis, we focus our attention on the case when $\mathcal{X}$ is a discrete (but possibly large depending exponentially on the optimization dimension $d$) set, which has size $|\mathcal{X}|$. As discussed earlier, to evaluate our algorithm we consider the criterion of regret. Let $x_{t,i}$ denote the query point of the $i$-th agent at the $t$-th time. Throughout, we use the notation $[N] := \{1,2,\dots,N\}$ (for any positive integer $N$).

We proceed to discuss some preliminaries in order to explain our algorithm. In the sequel, we denote $f^* := f(x^*)$. Let $X^{t,m} := \{ x_{1,1},\dots, x_{1,m},\dots, x_{t,1},\dots,x_{t,m}\} \in \mathcal{X}^{tm}$ denote the $tm$ points evaluated by the algorithm after $t$ iterations where $m$ points were evaluated per iteration, with $x_{\tau,j}$ denoting the $j$-th point evaluated at the $\tau$-th batch; for notational convenience, we omit the dependence on the batch number $m$ and refer to $X^{t,m}$ as $X^t$ througout the paper. Let $\bm{y}_t$ denotes $\{f(x') + \epsilon' \}_{x' \in X^t}$, where we recall that $\epsilon' \sim N(0,\sigma_n^2)$, and $\mathcal{F}_t:=\left\{X^t,\bm{y}_t\right\}$.
Given data $\mathcal{F}_t$, 
for any $x \in \mathcal{X}$, we note that $f \mid \mathcal{F}_t \sim \textup{GP}(\mu_t(x), k_t(x,x'))$, with
\begin{align*}
    &\mu_t(x) = \bm{k}_t(x)^\top (\bm{K}_t + \sigma_n^2 \bm{I})^{-1} \bm{y}_t, \quad k_t(x,x') = k(x,x') - \bm{k}_t(x)^\top(\bm{K}_t + \sigma_n^2\bm{I})^{-1}\bm{k}_t(x'),  
\end{align*}
where $\bm{K}_t := [k(x',x^{''})]_{x', x^{''} \in X^t}$ denotes the empirical kernel matrix, $\bm{k}_t(x) := [k(x',x)]_{x' \in X^t}$. In particular, for any $x \in \mathcal{X}$, we have that
$f(x) \mid \mathcal{F}_t \sim N(\mu_t(x), \sigma_t^2(x))$, where the posterior variance satisfies
\begin{align}
\label{eq:posterior_variance_defn}
\sigma_t^2(x) = k(x,x) - \bm{k}_t(x)^\top(\bm{K}_t + \sigma_n^2\bm{I})^{-1}\bm{k}_t(x).
\end{align}
For any set of $B$ points $\{x_b\}_{b \in [B]} \in \mathcal{X}$, we also find it useful to introduce the following notation of posterior variance $\sigma_t^2(x \mid \{x_b\}_{b \in [B]})$, where
\begin{align}
    & \ \sigma_t^2(x \mid \{x_b\}_{b \in [B]}) =   k(x,x) - \bm{k}_{t,B}(x)^\top(\bm{K}_{X^t \cup [B]} + \sigma_n^2\bm{I})^{-1}\bm{k}_{t,B}(x),
\end{align}
where $\bm{k}_{t,B}(x)$ represents the concatenation of $\bm{k}_t(x)$ and $[k(x_b,x)]_{b \in [B]}$, and $\bm{K}_{X^t \cup [B]} \in \bbR^{(tm+B)\times (tm+B)}$ is a block matrix of the form
\begin{align*}
    \bm{K}_{X^t \cup [B]} = \begin{bmatrix}
        \bm{K}_t & \bm{K}_{t,B} \\
        \bm{K}_{t,B}^\top  &\bm{K}_{B,B}
    \end{bmatrix},
\end{align*}
where $\bm{K}_{t,B} = [k(x',x_b)]_{x' \in X^t, b \in [B]} \in \bbR^{tm \times B}$, and $\bm{K}_{B,B} = [k(x_b, x_{b'})]_{b,b' \in [B]} \in \bbR^{B \times B}$. In other words, $\sigma_t^2(x \mid \{x_b\}_{b \in [B]})$ denotes the posterior variance conditional on having evaluated $X^t$ as well as an additional set of points $\{x_b\}_{b \in [B]}$.

 \vspace{-0.5em}
\section{Algorithm and statement of main result (Theorem \ref{theorem:main_result})}\label{sec:algorithm}
 \vspace{-0.25em}
For clarity, we first describe our algorithm in the case when the batch size $m$ is 1. At each time $t$, the algorithm chooses the next sample according to the following criterion:
\begin{align}
    \label{eq:alg_sampling_rule}
    x_{t+1} \in  \argmin_{x \in \mathcal{X}}  \frac{\tilde{f}_t^* - \mu_t(x) }{\sigma_t(x)} =: \Psi_t(x),
\end{align}
where $\tilde{f}_t^* := \max_x \tilde{f}_t(x)$ and$\tilde{f}_t$ is a single sample from the distribution $f \mid \mathcal{F}_t$. The numerator may be regarded as a TS approximation of the regret incurred by the action $x$,
whilst the denominator is the predictive standard deviation/uncertainty of the point $x$. This explains the name of our algorithm. In this case, the sampling scheme balances choosing points with high predictive mean with those which have high predictive uncertainty. 

In the batch setting, where we have to choose a batch of points simultaneously before receiving feedback, our algorithm takes the form 
\begin{align}
    & x_{t+1,1}^\alg \in \argmin_{x \in \mathcal{X}} \frac{\tilde{f}_{t,1}^* - \mu_t(x)}{\sigma_t(x)} \nonumber \\
    & x_{t+1,2}^\alg \in \argmin_{x \in \mathcal{X}} \frac{\tilde{f}_{t,2}^* - \mu_t(x)}{\sigma_t\left(x \mid \{x_{t+1,1}^\alg\}\right)} \nonumber \\
    & \qquad \qquad \qquad \vdots \nonumber \\
    & x_{t+1,m}^\alg \in \argmin_{x \in \mathcal{X}} \frac{\tilde{f}_{t,m}^* - \mu_t(x)}{\sigma_t\left(x \mid \{x_{t+1,j}^\alg \}_{j=1}^{m-1}\right)}\label{eq:alg_sampling_rule_batch}
\end{align}
where for each $i \in [m]$, $\tilde{f}_{t,i}^* := \max_x \tilde{f}_{t,i} (x)$, where $\tilde{f}_{t,i}$ denotes an independent sample from the distribution $f \mid \mathcal{F}_t$. Meanwhile, $\sigma_t(x \mid \{x_{t+1,j}\}_{j=1}^{\tau})$ denotes the predictive standard deviation of the posterior GP conditional on $\left\{X^t, \{ x_{t+1,j}\}_{j=1}^\tau\right\}$; 
we recall that the predictive variance only depends on the points that have been picked, and not the values of those points (see \eq{eq:posterior_variance_defn}). Intuitively, the denominator in \eq{eq:alg_sampling_rule_batch} encourages exploration, since it is large when the sample points are both uncertain conditional on the knowledge so far ($\mathcal{F}_t$) and are spaced far apart. Moreover, the numerator in \eq{eq:alg_sampling_rule_batch} is smaller for points with higher predictive means conditional on $\mathcal{F}_t$. We note that while our algorithm does not lend itself to parallelization of the batch selection process at each round, its computational complexity scales at each round only linearly with the number of points $m$ in a batch, which is significantly better than the exponential (in $m$) computational complexity in some other batch BO algorithms such as DPPTS~\cite{nava_diversified_2022}.


\begin{algorithm}[tb]
   \caption{$\alg$}
   \label{alg:main_alg}
\begin{algorithmic}[1]
   \STATE {\bfseries Input:} Input set $\mathcal{X}$; GP Prior $\mu_0  = 0$, $k$, output noise standard deviation $\sigma_n$; batch size $m$
   \FOR{$t=0, 1, \cdots, T-1$}
   \STATE Sample $m$ i.i.d copies of $\tilde{f}_{t,i} \sim f \mid \mathcal{F}_t$, and set $\tilde{f}_{t,i}^* = \max_x \tilde{f}_{t,i} (x)$.
   \STATE Choose 
    \begin{align*}
    & x_{t+1,1}^\alg \in \argmin_{x \in \mathcal{X}} \frac{\tilde{f}_{t,1}^* - \mu_t(x)}{\sigma_t(x)} \\
    & x_{t+1,2}^\alg \in \argmin_{x \in \mathcal{X}} \frac{\tilde{f}_{t,2}^* - \mu_t(x)}{\sigma_t(x \mid \{x_{t+1,1}^\alg\})} \\
    & \qquad \qquad \qquad  \vdots  \\
    & x_{t+1,m}^\alg \in \argmin_{x \in \mathcal{X}} \frac{\tilde{f}_{t,m}^* - \mu_t(x)}{\sigma_t(x \mid \{x_{t+1,j}^\alg \}_{j=1}^{m-1})}
    \end{align*}
    \STATE  Observe $y_{t+1,i} = f(x_{t+1,i}^\alg) + \epsilon_{t+1,i}$ for each $i \in [m]$
    \STATE  Perform Bayesian update to obtain $\mu_{t+1}, \sigma_{t+1}$
   \ENDFOR
\end{algorithmic}
\end{algorithm}

 

We proceed now to state the main result for the performance of our algorithm.

\begin{theorem}
\label{theorem:main_result}
Suppose $k(x,x') \leq 1$ for all $x,x'$. Let $\mathcal{X}$ be a discrete set, where $\abs{\mathcal{X}} \geq 2$. Then, running $\alg$  for $f\sim GP(0, k(\cdot, \cdot))$ will have the following regret,
\begin{align*}
    &\mathbb{E}\left[R_{T,m}\right] = O\left(\rho_{m} \sqrt{Tm \gamma_{Tm}}  \sqrt{\log\left( \abs*{\mathcal{X}} (Tm)^3\right)} \right), 
\end{align*}
which implies that the simple regret satisfies
\begin{align*}
    &\mathbb{E}\left[S_{T,m}\right] = O\left(\frac{\rho_{m} \sqrt{\gamma_{Tm}}}{\sqrt{Tm}}  \sqrt{\log\left( \abs*{\mathcal{X}} (Tm)^3\right)} \right)
\end{align*}
where $\rho_m := \max_{x \in \mathcal{X}, \tau, \tilde{A}_m \subset \mathcal{X}, \abs{\tilde{A}_m} \leq m} \frac{\sigma_\tau(x)}{\sigma_\tau(x \mid \tilde{A}_m)}$ denotes maximal decrease in posterior standard deviation resulting from conditioning on any additional set of samples $\tilde{A}_m$ of cardinality up to $m$, $\gamma_{Tm}$ denotes the maximal informational gain by observing $Tm$ elements, and the expectation is taken over the random draw of $f \sim GP(0, k(\cdot,\cdot))$ as well as the stochasticity of the measurement noises and the stochasticity of the TS draws.
\end{theorem}
\begin{proof}
    We defer the proof to Section \ref{sec:analysis} below.
\end{proof}
We compare now our results to existing theoretical bounds in the literature.\footnote{For clarity, in the sequel, we focus on comparing to results that assume that $f$ is drawn from a known GP (which is the setting we study), as opposed to the case when $f$ is drawn from a RKHS with bounded norm. This is to avoid confusion since in some of these works, in particular the papers on GP-UCB~\cite{srinivas2009gaussian} and BUCB~\cite{desautels2014parallelizing}, both scenarios are studied.} For sequential BO with a total of $Tm$ function evaluations, the simple regret for UCB \cite{srinivas2009gaussian} and Thompson Sampling \cite{kandasamy2018parallelised} is $\tilde{O}\left(\frac{\sqrt{\gamma_{Tm}}}{\sqrt{Tm}}\right)$. For batch BO, the simple regret for both BUCB \cite{desautels2014parallelizing} and parallel Thompson Sampling \cite{kandasamy2018parallelised} both scale as $\tilde{O}\left(\frac{\rho_m \sqrt{\gamma_{Tm}}}{\sqrt{Tm}}\right)$. Our regret matches the dependence of these two algorithms. This also implies that our algorithm performs nearly as well as a sequential algorithm with the same number of function evaluations $(Tm)$, up to a factor of $\rho_m$. In fact, it has been shown that with an appropriate initialization strategy detailed in \cite{desautels2014parallelizing}, the $\rho_m$ term can be driven down to $\tilde{O}(1)$, which implies then that batch BO can essentially achieve the same convergence rate as standard sequential BO. For completeness, we provide in Appendix \ref{appendix:rho_m_bound} a detailed discussion of the initialization strategy to achieve this reduction. For the $\gamma_{Tm}$ information gain quantity, well-known bounds exist in the literature for several commonly-used kernels (such as the linear, squared exponential and Matern kernels), which we state in Appendix \ref{appendix:gamma_bounds} for completeness. Finally, in Appendix \ref{appendix:end_to_end} we provide the convergence rate of our algorithm when combined with the initialization strategy in \cite{desautels2014parallelizing}. We note that results in Appendices \ref{appendix:rho_m_bound}, \ref{appendix:gamma_bounds} and \ref{appendix:end_to_end} are mainly statements/applications of known results to our setting, which is why we defer them to the appendix.

\begin{remark}
\label{remark:finite_D}
    We note that while our analysis focused on the discrete case, for kernels where the resulting GP sample functions are differentiable with high probability, such as the squared exponential kernel kernel or the Matern kernel (with $\nu$ parameter at least 1.5), the analysis of regret for a bounded compact set $\mathcal{X} \in \mathbb{R}^d$ can be essentially reduced to the analysis of a discretization $\bar{X}$ of $\mathcal{X}$ where $\abs*{\bar{X}} = O(\epsilon^{-d})$, where $0 < \epsilon < 1$ is a discretization parameter that is a function of the smoothness of the kernel; see for instance the analysis in \cite{srinivas2009gaussian}.
    In this case, the regret bound achieved by our algorithm is $\tilde{O}\left(\rho_m \sqrt{d \gamma_{Tm} Tm} \right)$. To compare this to other batch BO algorithms, take the BUCB algorithm as an example~\cite{desautels2014parallelizing}. It can be shown that for a smooth kernel and a bounded compact optimization set $\mathcal{X} \in \mathbb{R}^d$, BUCB also achieves a rate of  $\tilde{O}\left(\rho_m \sqrt{\beta_{Tm}\gamma_{Tm} Tm} \right)$, where $\beta_{Tm}$ term is a confidence parameter term and is of the order $\tilde{O}(d)$ (see Theorem 2 in \cite{desautels2014parallelizing}). This is the same rate achieved by our algorithm. This shows that without loss of generality, we may focus our analysis on the discrete case.
\end{remark}



 \vspace{-0.5em}
\section{Proof of Theorem \ref{theorem:main_result}}\label{sec:analysis}
 \vspace{-0.25em}
 
\paragraph{Decomposition of regret}
To provide a proof outline of Theorem \ref{theorem:main_result}, we first have the following result which decomposes $\xpt{R_{T,m}}$ into two quantities which we will proceed to bound later. 

\begin{lemma}
\label{lemma:xpt_R_R_tilde_decomp}
Let $R_{T,m} = \sum_{t=0}^{T-1} \sum_{i \in [m]} f^* - f(x_{t+1,i}^\alg)$, and $\tilde{R}_{T,m} = \sum_{t=0}^{T-1} \sum_{i \in [m]} \tilde{f}_{t,i}^* - f(x_{t+1,i}^\alg)$. Then, for any event $\mathcal{G}$, we have
\begin{align*}
    \mathbb{E}[R_{T,m}] = \xpt{\tilde{R}_{T,m}} = & \  \mathbb{E}[\tilde{R}_{T,m} 1_{\mathcal{G}}] + \mathbb{E}[\tilde{R}_{T,m} 1_{\mathcal{G}}^c] \leq  \mathbb{E}[\tilde{R}_{T,m} 1_{\mathcal{G}}] + \sqrt{\mathbb{E}[\tilde{R}_{T,m}^2] \mathbb{P}(\mathcal{G}^c)},
\end{align*}
\end{lemma}
\begin{proof}
We first observe that by the tower property, for any $t \in [T]$ and $i \in [m]$, we have
\begin{align*}
    \xpt{f^*} = \xpt{\xpt{\tilde{f}_{t,i}^* \mid \mathcal{F}_t}} = \xpt{\tilde{f}_{t,i}^*},
\end{align*}
where we recall that $\tilde{f}_{t,i}^* = \max_{x \in \mathcal{X}} \tilde{f}_{t,i}(x)$, and $\tilde{f}_{t,i}$ is a random draw (of the $i$-th agent at the $t$-th round) from $f \mid \mathcal{F}_t$.
Thus, 
\begin{align*}
    \xpt{R_{T,m}} = \xpt{\sum_{t=0}^{T-1} \sum_{i \in [m]} f^* - f(x_{t+1,i}^\alg)} = \xpt{\sum_{t=0}^{T-1} \sum_{i \in [m]} \tilde{f}_{t,i}^* - f(x_{t+1,i}^\alg)}.
\end{align*}
Letting $\tilde{R}_{T,m} := \sum_{t=0}^{T-1} \sum_{i \in [m]} \tilde{f}_{t,i}^* - f(x_{t+1,i}^\alg)$, we see then that $\xpt{R_{T,m}} = \xpt{\tilde{R}_{T,m}}$.\footnote{We  note that proving a regret bound in expectation allows us to relate $R_{T,m}$ to $\tilde{R}_{T,m}$, where the latter can then be bounded using our algorithmic choice of minimizing the TS regret-to-sigma ratio. Achieving a high-probability regret bound may be possible, but will likely involve further tweaks to the algorithm such as averaging more samples of $\tilde{f}_{t,i}^*$ to estimate the approximation to the true $f^*$ when sampling the $i$-th point in the $t$-th iteration.} Observe that for any event $\mathcal{G}$, we have
\begin{align*}
    \mathbb{E}[R_{T,m}] & \ = \xpt{\tilde{R}_{T,m}} =  \mathbb{E}[\tilde{R}_{T,m} 1_{\mathcal{G}}] + \mathbb{E}[\tilde{R}_{T,m} 1_{\mathcal{G}}^c] \\
    \leq & \ \mathbb{E}[\tilde{R}_{T,m} 1_{\mathcal{G}}] + \sqrt{\mathbb{E}[\tilde{R}_{T,m}^2] \xpt{(1_{\mathcal{G}}^c)^2}} \leq \mathbb{E}[\tilde{R}_{T,m} 1_{\mathcal{G}}] + \sqrt{\mathbb{E}[\tilde{R}_{T,m}^2] \mathbb{P}(\mathcal{G}^c)},
\end{align*}
where the first inequality follows from applying Cauchy-Schwarz to the term $\mathbb{E}[\tilde{R}_{T,m} 1_{\mathcal{G}}^c]$.
\end{proof}
\paragraph{Proof outline of Theorem \ref{theorem:main_result}} 
 Equipped with Lemma \ref{lemma:xpt_R_R_tilde_decomp}, we have the following roadmap to bounding the regret of our algorithm.
\begin{enumerate}[leftmargin=0.5cm]
     \item First, given any $0 < \delta < 1$, we define an event $\mathcal{G}(\delta)$ on which we have an almost sure bound on $\tilde{R}_{T,m} 1_{\mathcal{G}(\delta)}$ which translates to a bound on $\xpt{\tilde{R}_{T,m}1_{\mathcal{G}(\delta)}}$. Conceptually, $\mathcal{G}(\delta)$ can be thought of as a ``likely'' event that happens with probability at least $1 - O(\delta)$, on which $\tilde{R}_{T,m} 1_{\mathcal{G}(\delta)}$ can be shown to be bounded. Concretely, $\mathcal{G}(\delta)$ is the intersection 
     of two events $\mathcal{G}^{(1)}(\delta)$ and $\mathcal{G}^{(2)}(\delta)$, which will later be defined in (\ref{eq:G1_event_union_defn}) and (\ref{eq:G2_event_defn}) respectively. To bound $\tilde{R}_{T,m}1_{\mathcal{G}(\delta)}$, we have the following steps.
     \begin{enumerate}[leftmargin=0.5cm]
         \item In Section \ref{subsection:decomposition_of_tilde_R}, we provide a general bound for $\tilde{R}_{T,m}$, decomposing it as a sum of two terms, which are
         \footnotesize
         \begin{align*}
            & \ S_1 := \bar{\Psi} \sqrt{Tm\sum_{t=0}^{T-1} \sum_{i=1}^m \sigma_t^2(x_{t+1,i}^\alg \mid \{x_{t+1,j}^\alg\}_{j=1}^{i-1})}, \\
            & \ S_2 := \sum_{t=0}^{T-1} \left(\sum_{i=1}^m  (\mu_t(x_{t+1,i}^\alg) - f(x_{t+1,i}^\alg))\right)
         \end{align*}
         \normalsize
         where
         \footnotesize
         \begin{align}
             \bar{\Psi} := \max_{t=0,\dots,T-1} \left(\max_{i \in [m]} \frac{\tilde{f}_{t,i}^* - \mu_{t}(x_{t+1,i}^\alg)}{\sigma_t(x_{t+1,i}^\alg \mid \{x_{t+1,j}^\alg\}_{j=1}^{i-1})} \right). \label{eq:first_bar_psi_defn}
         \end{align}
         \normalsize
         \item In Section \ref{subsec:bound_bar_psi}, we bound the term $\bar{\Psi}$ on the event $\mathcal{G}^{(1)}(\delta)$. We note that our bound of the term $\bar{\Psi}$ is the key novelty in our proof. 
         
         \item In Section \ref{subsec:bound_sum_of_posterior_variances}, we bound the term  $\sqrt{Tm\sum_{t=0}^{T-1} \sum_{i=1}^m \sigma_t^2(x_{t+1,i}^\alg \mid \{x_{t+1,j}^\alg\}_{j=1}^{i-1})}$. 
         \item In Section \ref{subsec:bounding_sum_mu_minus_f}, on the event $\mathcal{G}^{(2)}(\delta)$, we bound the term $S_2$.
     \end{enumerate}
     \item In Section \ref{subsec:proof_main_result_wrap_up}, we first combine the bounds in the preceding step to yield a bound on $\tilde{R}_{T,m}1_{\mathcal{G}(\delta)}$. We then combine this with a bound on $\sqrt{\mathbb{E}[\tilde{R}_{T,m}^2] \mathbb{P}(\mathcal{G}^c)}$ to wrap up the proof of Theorem \ref{theorem:main_result}.
 \end{enumerate}


\subsection{Decomposition of $\tilde{R}_{T,m}$}
\label{subsection:decomposition_of_tilde_R}
The following helpful lemma demonstrates how we can decompose $\tilde{R}_{T,m}$ in terms of $\bar{\Psi}$ (defined in (\ref{eq:first_bar_psi_defn})), which we also refer to as the Regret-to-Sigma ratio (RSR), and two other quantities, namely 
\footnotesize
$$\sum_{t=0}^{T-1} \left(\sum_{i=1}^m  (\mu_t(x_{t+1,i}^\alg) - f(x_{t+1,i}^\alg))\right), \quad   \sqrt{Tm\sum_{t=0}^{T-1} \sum_{i=1}^m \sigma_t^2(x_{t+1,i}^\alg \mid \{x_{t+1,j}^\alg\}_{j=1}^{i-1})}.$$ 
\normalsize
The term $\bar{\Psi}$ will be bounded in Section \ref{subsec:bound_bar_psi} and the term \footnotesize $\sqrt{Tm\sum_{t=0}^{T-1} \sum_{i=1}^m \sigma_t^2(x_{t+1,i}^\alg \mid \{x_{t+1,j}^\alg\}_{j=1}^{i-1})}$ \normalsize will be bounded in Section \ref{subsec:bound_sum_of_posterior_variances}. Meanwhile, the term \footnotesize $\sum_{t=0}^{T-1} \left(\sum_{i=1}^m  (\mu_t(x_{t+1,i}^\alg) - f(x_{t+1,i}^\alg))\right)$ \normalsize will be bounded in Section \ref{subsec:bounding_sum_mu_minus_f}.

\begin{lemma}
\small
\label{lemma:R_tilde_detailed_decomp}
    The term $\tilde{R}_{T,m}$ can be decomposed as
    \begin{align}
        \tilde{R}_{T,m} = & \ \sum_{t=0}^{T-1} \sum_{i=1}^m \tilde{f}_{t,i}^* - f(x_{t+1,i}^{\alg}) \nonumber \\ 
        \leq & \ \sum_{t=0}^{T-1} \left(\sum_{i=1}^m  (\mu_t(x_{t+1,i}^\alg) - f(x_{t+1,i}^\alg))\right) \nonumber \\
        \quad & \ +  \bar{\Psi} \sqrt{Tm\sum_{t=0}^{T-1} \sum_{i=1}^m \sigma_t^2(x_{t+1,i}^\alg \mid \{x_{t+1,j}^\alg\}_{j=1}^{i-1})} \label{eq:part1_outline_result},
    \end{align}
\normalsize
where 
\small
\begin{align*}
    \bar{\Psi} := \max_{t=0,\dots,T-1} \left(\max_{i \in [m]} \frac{\tilde{f}_{t,i}^* - \mu_{t}(x_{t+1,i}^\alg)}{\sigma_t(x_{t+1,i}^\alg \mid \{x_{t+1,j}^\alg\}_{j=1}^{i-1})} \right).
\end{align*}
\normalsize
\end{lemma}
\begin{proof}
    We observe that 
    \small
    \begin{align*}
        \tilde{R}_{T,m} = & \ \sum_{t=0}^{T-1} \sum_{i=1}^m \tilde{f}_{t,i}^* - f(x_{t+1,i}^{\alg}) \nonumber \\
        = & \ \sum_{t=0}^{T-1} \sum_{i=1}^m \tilde{f}_{t,i}^* - \mu_t(x_{t+1,i}^{\alg}) + \mu_t(x_{t+1,i}^{\alg}) - f(x_{t+1,i}^{\alg}) \nonumber \\
        = & \ \sum_{t=0}^{T-1} \left(\sum_{i=1}^m  (\mu_t(x_{t+1,i}^\alg) - f(x_{t+1,i}^\alg))\right) \nonumber \\
        \quad \ \  & + \sum_{t=0}^{T-1} \sum_{i=1}^m \frac{\left(\!\tilde{f}_{t,i}^* \!-\! \mu_t(x_{t+1,i}^{\alg})\!\right) \sigma_t(x_{t+1,i}^\alg \!\mid\! \{x_{t+1,j}^\alg\}_{j=1}^{i-1}) }{ \sigma_t(x_{t+1,i}^\alg \mid \{x_{t+1,j}^\alg\}_{j=1}^{i-1})}   \nonumber \\
        \labelrel\leq{eq:use_RSR_bdd} & \ \sum_{t=0}^{T-1} \left(\sum_{i=1}^m (\mu_t(x_{t+1,i}^\alg) - f(x_{t+1,i}^\alg))\right) \nonumber \\
        \quad \ \ & + \bar{\Psi} \sum_{t=0}^{T-1} \sum_{i=1}^m \sigma_t(x_{t+1,i}^\alg \mid \{x_{t+1,j}^\alg\}_{j=1}^{i-1}) \nonumber \\
\labelrel\leq{eq:use_Cauchy_bdd_sqrt_sum} & \ \sum_{t=0}^{T-1} \left(\sum_{i=1}^m  (\mu_t(x_{t+1,i}^\alg) - f(x_{t+1,i}^\alg))\right) \nonumber \\
\quad \ \ & + \bar{\Psi} \sqrt{Tm\sum_{t=0}^{T-1} \sum_{i=1}^m \sigma_t^2(x_{t+1,i}^\alg \mid \{x_{t+1,j}^\alg\}_{j=1}^{i-1})}
    \end{align*}
    \normalsize
    Above, in obtaining (\ref{eq:use_RSR_bdd}), we define the maximum Regret-to-Sigma Ratio (RSR) encountered during the course of the algorithm as
    \footnotesize
    $\bar{\Psi} := \max_{0 \leq t \leq T-1, i \in [m]} \frac{\tilde{f}_{t,i}^* - \mu_{t}(x_{t+1,i}^\alg)}{\sigma_t(x_{t+1,i}^\alg \mid \{x_{t+1,j}^\alg\}_{j=1}^{i-1})}.$ 
    \normalsize
    In addition, we used Cauchy-Schwarz to derive (\ref{eq:use_Cauchy_bdd_sqrt_sum}). This completes our proof.
\end{proof}

\subsection{Bounding $\bar{\Psi}$}
\label{subsec:bound_bar_psi} 
Here, we bound  $\bar{\Psi}$, defined in (\ref{eq:first_bar_psi_defn}), which represents the maximum Regret-to-Sigma Ratio (RSR) encountered during the course of the algorithm. The term $\bar{\Psi}$ appeared in the decomposition of the regret term $\tilde{R}_{T,m}$ in (\ref{eq:part1_outline_result}). In Lemma \ref{lemma:bar_Psi_bdd} to appear later, we will bound $\bar{\Psi}$ by using a probabilistic argument to show that $\bar{\Psi}$ is always bounded on a ``likely'' event $\mathcal{G}^{(1)}(\delta)$ (which will be defined in (\ref{eq:G1_event_union_defn})) which happens with probability at least $1-\delta$. Before we state and prove Lemma \ref{lemma:bar_Psi_bdd}, we first show the following key technical result (Lemma \ref{lemma:max-gaussian-ratio-bdd}), which bounds the RSR for a collection of Gaussian variables.

\begin{lemma}
    \label{lemma:max-gaussian-ratio-bdd}
    Suppose $\bm{Y} \sim N(\bm{\mu}, \bm{\Sigma})$, where $\bm{\mu} \in \mathbb{R}^D$ and $\bm{\Sigma} \succ \bm{0}_{D \times D}$. For each $j \in [D]$, we denote $\sigma_j^2 := \bm{\Sigma}_{j,j}$. Let $\ell^* = \argmax_{j \in [D]} \bm{Y}_j$, and denote $\bm{Y}^* = \max_{j \in [D]} \bm{Y}_j = \bm{Y}_{\ell^*}$. For any $\delta > 0$, define the event
    \small
\begin{align}
        \label{eq:all_ratio_bdd_event}
        \mathcal{E}(\delta) := \left\{\forall \ell \in [D]: \abs*{\frac{\bm{Y}_\ell -\mu_\ell}{\sigma_\ell}} \leq \sqrt{2\log(D/\delta)}\right\}.
    \end{align}
    \normalsize
    Then, this event happens with probability at least $1 - \delta$. 
    Moreover, on this event, we have
    \small
    $$\min_{\ell \in [D]} \frac{\bm{Y}^* - \bm{\mu}_{\ell}}{\sigma_{\ell}} \leq \frac{\bm{Y}^* - \bm{\mu}_{\ell^*}}{\sigma_{\ell^*}} \leq \sqrt{2\log(D/\delta)} $$
    \normalsize
\end{lemma}
\begin{proof}
    Note that by a standard subGaussian concentration bound, for each $\ell \in [D]$, for any $t > 0$,
    \begin{align*}
        P\left(\abs*{\frac{\bm{Y}_\ell -\mu_\ell}{\sigma_\ell}} \geq t\right) \leq 2\exp(-t^2/2) 
    \end{align*}
    Pick $t = \sqrt{2\log(2D/\delta)}$. Then, it follows that for any $\ell \in [D]$,
    \begin{align*}
        P\left(\abs*{\frac{\bm{Y}_\ell -\mu_\ell}{\sigma_\ell}}\geq \sqrt{2\log(D/\delta)}\right) \leq & \  2\exp\left(-\frac{(\sqrt{2\log(2D/\delta)})^2}{2}\right) = \frac{\delta}{D}.
    \end{align*}
    Thus, by applying union bound, we have that 
    \begin{align}
    \label{eq:all_ratio_high_prob_bdd}
        P\left(\forall \ell \in [D]: \abs*{\frac{\bm{Y}_\ell -\mu_\ell}{\sigma_\ell}} \leq \sqrt{2\log(2D/\delta)}\right) \geq 1 - \delta.
    \end{align}
    Consider $\ell^*$ such that $\bm{Y}_{\ell^*} = \max_{\ell \in [D]} \bm{Y}_\ell$. Then, it follows by \eq{eq:all_ratio_high_prob_bdd} that 
    \begin{align*}
        \min_{\ell \in [D]} \frac{\bm{Y}^* - \bm{\mu}_{\ell}}{\sigma_{\ell}} \leq \frac{\bm{Y}_{\ell^*} -\mu_{\ell^*}}{\sigma_{\ell^*}} \leq \sqrt{2\log(2D/\delta)} 
    \end{align*}
    also holds with probability at least $1 - \delta$.
\end{proof}

We are now ready to state and prove Lemma \ref{lemma:bar_Psi_bdd}, which provides our bound on $\bar{\Psi}$. 

\begin{lemma}
\label{lemma:bar_Psi_bdd}
Define the events
\small
 \begin{align}
 \label{eq:G1_ti_event_union_defn}
    &\mathcal{G}_{t,i}^{(1)}(\delta) := \left\{\forall x \in \mathcal{X}: \abs*{\frac{\tilde{f}_{t,i}(x) -\mu_t(x)}{\sigma_t(x)}} \leq \sqrt{2\log(2\abs*{\mathcal{X}}mT/\delta)}\right\}.
 \end{align}
\normalsize
 Define also the union of these events across all rounds $t \in [T]$ and each index $i \in [m]$ in the batch
\small
 \begin{align}
 \label{eq:G1_event_union_defn}
    &\mathcal{G}^{(1)}(\delta) := \bigcup_{0 \leq t \leq T-1, i \in [m]} \mathcal{G}_{t,i}^{(1)}(\delta).
 \end{align}
 \normalsize
 Then, the event $\mathcal{G}^{(1)}(\delta)$ happens with probability at least $1 - \delta$. Moreover, on this event, we have
 \small
    \begin{align}
 \bar{\Psi} = & \ \max_{0 \leq t \leq T-1, i \in [m]} \Psi_{t,i}(x_{t+1,i}^{\alg}) \nonumber \\
= & \  \max_{0 \leq t \leq T-1, i \in [m]} \frac{\tilde{f}_{t,i}^* - \mu_t(x_{t+1,i}^\alg)}{ \sigma_t(x_{t+1,i}^\alg \mid \{x_{t+1,j}^\alg\}_{j=1}^{i-1})}  \leq \sqrt{2\log (2\abs*{\mathcal{X}}mT/\delta)} 
    \rho_m \label{eq:RSR_bdd},
    \end{align}
    \normalsize
    where 
    \small
    \begin{align*}
        \Psi_{t,i}(x) := \frac{\tilde{f}_{t,i}^* - \mu_t(x)}{ \sigma_t(x \mid \{x_{t+1,j}^{\alg} \}_{j=1}^{i-1})},
    \end{align*}
    \normalsize
    and
    $\rho_m := \max_{x \in \mathcal{X}, \tau, \tilde{A} \subset \mathcal{X}, \abs{\tilde{A}} \leq m} \frac{\sigma_\tau(x)}{\sigma_\tau(x \mid \tilde{A}_m)}$
    denotes the maximal decrease in posterior variance resulting from conditioning on an additional set of samples $\tilde{A}_m$ of cardinality up to $m$. 
\end{lemma}
\begin{proof}
We start by noting that at any time $t$, that for each $i \in [m]$, $\tilde{f}_{t,i}^* := \max_x \tilde{f}_{t,i}(x)$, where $\tilde{f}_{t,i}$ is an independent sample from $f \mid \mathcal{F}_t$. Let $x_{t+1,i}^{\ts} := \argmax_x \tilde{f}_{t,i}(x)$; we use $\ts$ in the superscript of $x_{t+1,i}^{\ts}$ to represent the fact that if we performed Thompson sampling and drew $m$ independent samples from $x^* \mid \mathcal{F}_t$ to be our action, we will play exactly the policy $\{x_{t+1,i}^{\ts}\}_{i=1}^m$. By applying Lemma \ref{lemma:max-gaussian-ratio-bdd}, we see that for any $\delta > 0$, for a given $0 \leq t \leq T-1$ and $i \in [m]$, the event 
 \small
 \begin{align*}
\mathcal{G}_{t,i}^{(1)}(\delta) := \left\{\forall x \in \mathcal{X}: \abs*{\frac{\tilde{f}_{t,i}(x) -\mu_t(x)}{\sigma_t(x)}} \leq \sqrt{2\log(2\abs*{\mathcal{X}}mT/\delta)}\right\},
 \end{align*}
 \normalsize
happens with probability at least $1 - \delta/(Tm)$. Moreover, again using Lemma \ref{lemma:max-gaussian-ratio-bdd}, on this event, we have
\small
\begin{align*}
     \min_{x \in \mathcal{X}} \frac{\tilde{f}_{t,i}^* - \mu_t(x)}{ \sigma_t(x \mid \{x_{t+1,j}^\alg\}_{j=1}^{i-1})} \leq \frac{\left(\tilde{f}_{t,i}^* - \mu_t(x_{t+1,i}^{\ts})\right)}{\sigma_t(x_{t+1,i}^{\ts})} \leq \sqrt{2\log (2\abs*{\mathcal{X}}mT/\delta)}.
\end{align*}
\normalsize
    By denoting $\rho_m$ to be
    \small
    \begin{align}
    \label{eq:sigma_ratio_m_upper_bound}
        \rho_m := \max_{x \in \mathcal{X}} \max_{\tau} \max_{\tilde{A}_m \subset \mathcal{X}, \abs{\tilde{A}_m} \leq m}\frac{\sigma_\tau(x)}{\sigma_\tau(x \mid \tilde{A}_m)},
    \end{align}
    \normalsize
    we then obtain that 
    \small
    \begin{align*}
         \sigma_t(x_{t+1,i}^\ts)\leq \rho_m \sigma_t(x_{t+1,i}^\ts \mid \{x_{t+1,j}^\alg\}_{j=1}^{i-1}),  
    \end{align*}
    \normalsize
    which implies that on the event $\mathcal{G}_{t,i}^{(1)}(\delta)$,
    \small
    \begin{align*}
    \frac{\left(\tilde{f}_{t,i}^* - \mu_t(x_{t+1,i}^{\ts})\right)}{\sigma_t(x_{t+1,i}^{\ts} \mid \{x_{t+1,j}^\alg\}_{j=1}^{i-1})} \leq \sqrt{2\log (2\abs*{\mathcal{X}}mT/\delta)} \rho_m. 
    \end{align*}
    \normalsize
    Since 
    \small
    $$x_{t+1,i}^\alg \in \argmin_{x \in \mathcal{X}} \frac{\tilde{f}_{t,i}^* - \mu_t(x)}{ \sigma_t(x \mid \{x_{t+1,j}^\alg\}_{j=1}^{i-1})},$$
    \normalsize
    this implies that on the event $\mathcal{G}_{t,i}^{(1)}(\delta)$, which happens with probability at least $1 - \delta/(mT)$, we have
    \small
    \begin{align*}
& \ \frac{\tilde{f}_{t,i}^* - \mu_t(x_{t+1,i}^\alg)}{ \sigma_t(x_{t+1,i}^\alg \mid \{x_{t+1,j}^\alg\}_{j=1}^{i-1})} \leq  \frac{ \tilde{f}_{t,i}^* - \mu_t(x_{t+1,i}^\ts)}{\sigma_t(x_{t+1,i}^\ts \mid \{x_{t+1,j}^\alg\}_{j=1}^{i-1})} \leq   \sqrt{2\log (2\abs*{\mathcal{X}}mT/\delta)} \rho_m. 
    \end{align*}
    \normalsize
The final result then follows by a union bound over $0 \leq t \leq T-1$ and $i \in [m]$.
\end{proof}

\subsection{Bounding $\sqrt{Tm\sum_{t=0}^{T-1}\sum_{i=1}^m \sigma_t^2(x_{t+1,i}^\alg \mid \{x_{t+1,j}^\alg\}_{j=1}^{i-1})}$}
\label{subsec:bound_sum_of_posterior_variances}
In this subsection, we bound the term $\sqrt{Tm\sum_{t=0}^{T-1}\sum_{i=1}^m \sigma_t^2(x_{t+1,i}^\alg \mid \{x_{t+1,j}^\alg\}_{j=1}^{i-1})}$, which appeared in the decomposition of the regret term $\tilde{R}_{T,m}$ in (\ref{eq:part1_outline_result}).

\begin{lemma}
        \label{lemma:sum_sigma_informational bound}
        Suppose $k(x,x) \leq 1$ for each $x \in \mathcal{X}$. Then, letting $C_1 := 2\sigma_n^{-2}/\log(1 + \sigma_n^{-2})$, we have
        \small
        \begin{align*}
            \sqrt{Tm\sum_{t=0}^{T-1} \sum_{i=1}^m \sigma_t^2(x_{t+1,i}^\alg \mid \{x_{t+1,j}^\alg\}_{j=1}^{i-1})} \leq \sqrt{Tm\sigma_n^2 C_1 \gamma_{Tm}}, 
        \end{align*}
        \normalsize
        where (recall $I(X;Y)$ denotes the mutual information between any two random variables $X$ and $Y$)
        \small
        \begin{align*}
            \gamma_{Tm} := \sup_{A \subset \mathcal{X}, \abs{A} = Tm} I(\bm{y}_A; f_A).
        \end{align*}
        \normalsize
    \end{lemma}
    \begin{proof}
The proof follows by the calculations in Lemma 5.4 of \cite{srinivas2009gaussian}, and we restate them here for completeness. For notational simplicity, denote $\sigma_{t,i}^2 := \sigma_t^2(x_{t+1,i}^\alg \mid \{x_{t+1,j}^\alg\}_{j=1}^{i-1})$. Then, observe that since $\sigma_{t,i}^2 \leq 1$ (which follows from our initial assumption on $k$), we have that $\sigma_n^{-2} \sigma_{t,i}^2 \leq \sigma_n^{-2}$. Since $\frac{x}{\log (1 +x)}$ is an increasing function for $x > 0$, this implies then that
\footnotesize
$\frac{\sigma_n^{-2} \sigma_{t,i}^2}{\log(1 + \sigma_n^{-2} \sigma_{t,i}^2)} \leq \frac{\sigma_n^{-2}}{\log(1 + \sigma_n^{-2})}.$
\normalsize
Hence, 
\footnotesize
$$\sigma_{t,i}^2 =  \sigma_n^2 \left(\sigma_n^{-2} \sigma_{t,i}^2 \right) \leq \sigma_n^2 \left( \frac{2\sigma_n^{-2}}{\log(1 + \sigma_n^{-2})} \right) \frac{1}{2} \log(1 + \sigma_n^{-2} \sigma_{t,i}^2) = \sigma_n^2 C_1 \left(\frac{1}{2} \log(1 + \sigma_n^{-2} \sigma_{t,i}^2)\right).$$
\normalsize
Summing across $t$ and $i$, we thus have
\footnotesize
\begin{align*}
    \sum_{t=0}^{T-1} \sum_{i=1}^m \sigma_t^2(x_{t+1,i}^\alg \mid \{x_{t+1,j}^\alg\}_{j=1}^{i-1}) \leq & \ \sigma_n^2 C_1 \left(\sum_{t=0}^{T-1} \sum_{i=1}^m \frac{1}{2} \log(1 + \sigma_n^{-2} \sigma_t^2(x_{t+1,i}^\alg \mid \{x_{t+1,j}^\alg\}_{j=1}^{i-1})\right) \\
    = & \ \sigma_n^2 C_1 I(f; \bm{y}_{[Tm]}) \leq  \sigma_n^2 C_1 \gamma_{Tm},
\end{align*}
\normalsize
where the second last equality follows from Lemma \ref{lemma:info_gain_sigma_rship} in Appendix \ref{appendix:information_theory}, and the last inequality follows by definition of $\gamma_{Tm}$.
\end{proof}

\subsection{Bounding $\sum_{t=0}^{T-1} \sum_{i=1}^m \mu_t(x_{t+1,i}^\alg) - f(x_{t+1,i}^\alg)$}
\label{subsec:bounding_sum_mu_minus_f}

Here, we show that on an event $\mathcal{G}^{(2)}(\delta)$ which happens with probability at least $1- \delta$, the term $\sum_{t=0}^{T-1} \sum_{i=1}^m \mu_t(x_{t+1,i}^\alg) - f(x_{t+1,i}^\alg)$ can be bounded; this term appeared in the decomposition of $\tilde{R}_{T,m}$ in (\ref{eq:part1_outline_result}) of Lemma \ref{lemma:R_tilde_detailed_decomp}.
\begin{lemma}
\label{lemma:mu_minus_f_sum_bdd}
    Define the event
    \begin{align}
        \label{eq:G2_event_defn}
        \mathcal{G}^{(2)}(\delta) := \left\{\forall 0 \leq t \leq T-1, \forall x \in \mathcal{X}: \abs*{\mu_t(x) - f(x)} \leq \sqrt{2\log(2\abs*{\mathcal{X}}T/\delta)} \sigma_t(x) \right\}.
    \end{align}
    Then, this event happens with probability at least $1 - \delta$. Moreover, on this event, we have 
        \begin{align}
        & \ \sum_{t=0}^{T-1} \sum_{i=1}^m \mu_t(x_{t+1,i}^\alg) - f(x_{t+1,i}^\alg) \leq \sqrt{2\log(2\abs*{\mathcal{X}}T/\delta)}  \rho_m \sqrt{Tm\sigma_n^2 C_1 \gamma_{Tm}}, \label{eq:mu_minus_f_sum_bdd}
    \end{align}
\end{lemma}
\begin{proof}
Fix some $0 < \delta < 1$. Consider any $0 \leq t \leq T-1$. Then, for any $x \in \mathcal{X}$, since $f(x) \mid \mathcal{F}_t$ is a Gaussian random variable with mean $\mu_t(x)$ and standard deviation $\sigma_t(x)$, by applying a standard subGaussian concentration inequality (cf. the argument in Lemma \ref{lemma:max-gaussian-ratio-bdd}), with probability at least $1 - \delta/(\abs*{\mathcal{X}}T)$, we have
    $$\abs*{\mu_t(x) - f(x)} \leq \sqrt{2\log(2\abs*{\mathcal{X}}T/\delta)}.$$
    Taking a union bound over all $x \in \mathcal{X}$ and $0 \leq t \leq T-1$, the event
    \begin{align*}
        \mathcal{G}^{(2)}(\delta) := \left\{\forall 0 \leq t \leq T-1, \forall x \in \mathcal{X}: \abs*{\mu_t(x) - f(x)} \leq \sqrt{2\log(2\abs*{\mathcal{X}}T/\delta)} \sigma_t(x) \right\}
    \end{align*}
    happens with probability at least $1 - \delta$.
    Recalling the definition of $\rho_m$ as
    \begin{align}
    \label{eq:sigma_ratio_m_upper_bound_1st_time}
        \rho_m := \max_{x \in \mathcal{X}} \max_{\tau} \max_{\tilde{A}_m \subset \mathcal{X}, \abs{\tilde{A}_m} \leq m}\frac{\sigma_\tau(x)}{\sigma_\tau(x \mid \tilde{A}_m)},
    \end{align}
    it follows that on the event $\mathcal{G}^{(2)}(\delta)$, we have that for each $0 \leq t \leq T-1$ and $i \in [m]$, we have
    \begin{align*}
        \mu_t(x_{t+1,i}^\alg) - f(x_{t+1,i}^\alg) \leq & \  \sqrt{2\log(2\abs*{\mathcal{X}}T/\delta)} \sigma_t(x_{t+1,i}^\alg) \\
        \leq & \ \sqrt{2\log(2\abs*{\mathcal{X}}T/\delta)} \sigma_t(x_{t+1,i}^\alg \mid \{x_{t+1,i}^\alg\}_{j=1}^{i-1}) \rho_m.
    \end{align*}
    Thus, on the event $\mathcal{G}^{(2)}(\delta)$, we have
    \begin{align}
        & \ \sum_{t=0}^{T-1} \sum_{i=1}^m \mu_t(x_{t+1,i}^\alg) - f(x_{t+1,i}^\alg) \nonumber \\
        \leq & \ \sum_{t=0}^{T-1} \sum_{i=1}^m \sqrt{2\log(2\abs*{\mathcal{X}}T/\delta)} \sigma_t(x_{t+1,i}^\alg \mid \{x_{t+1,i}^\alg\}_{j=1}^{i-1}) \rho_m \nonumber \\
        \leq & \ \sqrt{2\log(2\abs*{\mathcal{X}}T/\delta)}  \rho_m \sqrt{Tm \sum_{t=0}^{T-1} \sum_{i=1}^m \sigma_t^2(x_{t+1,i}^\alg \mid \{x_{t+1,i}^\alg\}_{j=1}^{i-1})} \nonumber \\
        \leq & \ \sqrt{2\log(2\abs*{\mathcal{X}}T/\delta)}  \rho_m \sqrt{Tm\sigma_n^2 C_1 \gamma_{Tm}}, \nonumber
    \end{align}
    where the second-to-last inequality follows by Cauchy-Schwarz, and the final inequality uses the information-theoretic bound in Lemma \ref{lemma:sum_sigma_informational bound}.
\end{proof}

\subsection{Proof of Theorem \ref{theorem:main_result}}
\label{subsec:proof_main_result_wrap_up}

We are now ready to prove our main result, Theorem \ref{theorem:main_result}. 
\begin{proof}[Proof of Theorem \ref{theorem:main_result}]

Recall that by the derivations in Lemma \ref{lemma:R_tilde_detailed_decomp}, we have from (\ref{eq:part1_outline_result}) that
    \begin{align*}
        \tilde{R}_{T,m} = & \  \sum_{t=0}^{T-1} \sum_{i=1}^m \tilde{f}_{t,i}^* - f(x_{t+_1,i}^\alg) \\
        \leq & \ \sum_{t=0}^{T-1} \sum_{i=1}^m  \mu_t(x_{t+1,i}^\alg) - f(x_{t+1,i}^\alg) + \bar{\Psi} \sqrt{Tm\sum_{t=0}^{T-1} \sum_{i=1}^m \sigma_t^2(x_{t+1,i}^\alg \mid \{x_{t+1,j}^\alg\}_{j=1}^{i-1})}.
    \end{align*}
    Consider a fixed $0 < \delta < 1$. By combining 
    \begin{enumerate}[label=(\alph*)]
        \item the bound for $\bar{\Psi}$ in (\ref{eq:RSR_bdd}) of Lemma \ref{lemma:bar_Psi_bdd} on the event $\mathcal{G}^{(1)}(\delta)$,
        \item the bound for $\sqrt{Tm\sum_{t=0}^{T-1} \sum_{i=1}^m \sigma_t^2(x_{t+1,i}^\alg \mid \{x_{t+1,j}^\alg\}_{j=1}^{i-1})}$ in Lemma \ref{lemma:sum_sigma_informational bound}, 
        \item the bound for $\sum_{t=0}^{T-1} \sum_{i=1}^m  \mu_t(x_{t+1,i}^\alg) - f(x_{t+1,i}^\alg)$ on event $\mathcal{G}^{(2)}(\delta)$ in (\ref{eq:mu_minus_f_sum_bdd}) of Lemma \ref{lemma:mu_minus_f_sum_bdd},
    \end{enumerate}
    we obtain that (i) the event $ \mathcal{G}(\delta) := \mathcal{G}^{(1)}(\delta) \cap \mathcal{G}^{(2)}(\delta)$ (where $\mathcal{G}^{(1)}(\delta)$ is defined in (\ref{eq:G1_event_union_defn} and $\mathcal{G}^{(2)}(\delta)$ is defined in (\ref{eq:G2_event_defn})) happens with probability at least $1 - 2\delta$ and (ii) on this event, we have 
    \begin{align*}
        \tilde{R}_{T,m} = & \  \sum_{t=0}^{T-1} \sum_{i=1}^m \tilde{f}_{t,i}^* - f(x_{t+_1,i}^\alg)  \leq   2\sqrt{2\log(2\abs*{\mathcal{X}}mT/\delta)}  \rho_m \sqrt{Tm\sigma_n^2 C_1 \gamma_{Tm}}.
    \end{align*}
    Pick now $\delta_0 = (Tm)^{-2}/2$, and define the event $\mathcal{G} = \mathcal{G}(\delta_0)$. Note that
    \begin{align}
    \label{eq:prob_G_complement_bdd}
        \mathbb{P}\left(\mathcal{G}^c\right) \leq (Tm)^{-2}.
    \end{align}
    Since 
    \begin{align*}
       \tilde{R}_{T,m} 1_{\mathcal{G}} \leq & \  2\sqrt{2\log(2\abs*{\mathcal{X}}mT/\delta_0)}  \rho_m \sqrt{Tm\sigma_n^2 C_1 \gamma_{Tm}} \leq  2\sqrt{2\log(4\abs*{\mathcal{X}}(mT)^3}  \rho_m \sqrt{Tm\sigma_n^2 C_1 \gamma_{Tm}},
    \end{align*}
    it follows that $\xpt{\tilde{R}_{T,m}1_{\mathcal{G}}} \leq 2\sqrt{2\log(4\abs*{\mathcal{X}}(mT)^3}  \rho_m \sqrt{Tm\sigma_n^2 C_1 \gamma_{Tm}}$. Meanwhile, observe that
    \begin{align*}
        \xpt{\tilde{R}_{T,m} 1_{\mathcal{G}^c}} \leq \sqrt{\xpt{\tilde{R}_{T,m}^2} \mathbb{P}(\mathcal{G}^c)} \leq \sqrt{24\log \abs*{\mathcal{X}} (Tm)^3} \sqrt{(Tm)^{-2}} \leq \sqrt{24\log \abs*{\mathcal{X}} Tm},
    \end{align*}
    where the second-to-last inequality follows from a bound on $\xpt{\tilde{R}_{T,m}^2}$ in Lemma \ref{lemma:R_tilde_^2_bdd} which we state and prove in Appendix \ref{appendix:bound_xpt_tilde_R_sq} and the bound on $\mathbb{P}(\mathcal{G}^c)$ in (\ref{eq:prob_G_complement_bdd}). Thus, it follows from Lemma \ref{lemma:xpt_R_R_tilde_decomp} that
    \begin{align*}
        \xpt{R_{T,m}}  = \xpt{\tilde{R}_{T,m}} \leq & \ \xpt{\tilde{R}_{T,m}1_{\mathcal{G}}} +  \sqrt{\xpt{\tilde{R}_{T,m}^2} \mathbb{P}(\mathcal{G}^c)} \\
        \leq & \ 2\sqrt{2\log(4\abs*{\mathcal{X}}(mT)^3}  \rho_m \sqrt{Tm\sigma_n^2 C_1 \gamma_{Tm}} + \sqrt{24\log \abs*{\mathcal{X}} Tm}.
    \end{align*}
    This completes our proof. 
\end{proof}

 \vspace{-0.5em}
\section{Numerical results}\label{sec:simulations}
 \vspace{-0.25em}





The performance of our algorithm is compared against the following competitors: namely Batch UCB (BUCB, \cite{desautels2014parallelizing}), Thompson Sampling (TS, \cite{kandasamy2018parallelised}), GP-UCB with pure exploitation (UCBPE, \cite{contal2013parallel}),  Fully Distributed Bayesian Optimization with Stochastic Policies (SP, \cite{garcia2019fully}), a sequential kriging version of Expected Improvement (qEI, \cite{zhan2020expected}, \cite{hunt2020batch}, \cite{ginsbourger2008multi}), and  DPPTS~\cite{nava_diversified_2022} (which is a state-of-the-art batch variant of Thompson Sampling).

\subsection{Functions sampled from GP prior}
\begin{figure*}[h]
  \centering
  \includegraphics[width=.45\textwidth]{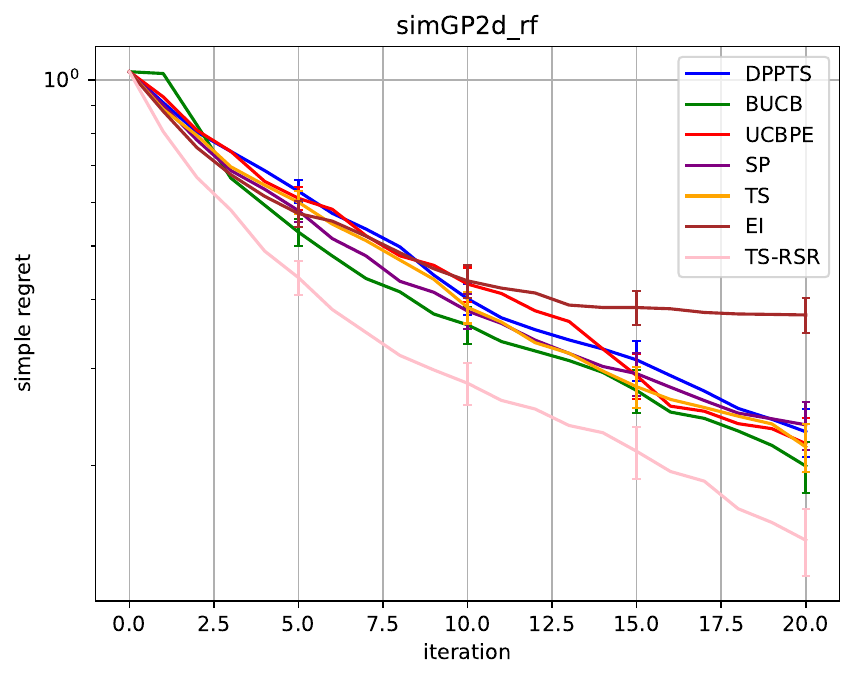}\hfill
\includegraphics[width=.45\textwidth]{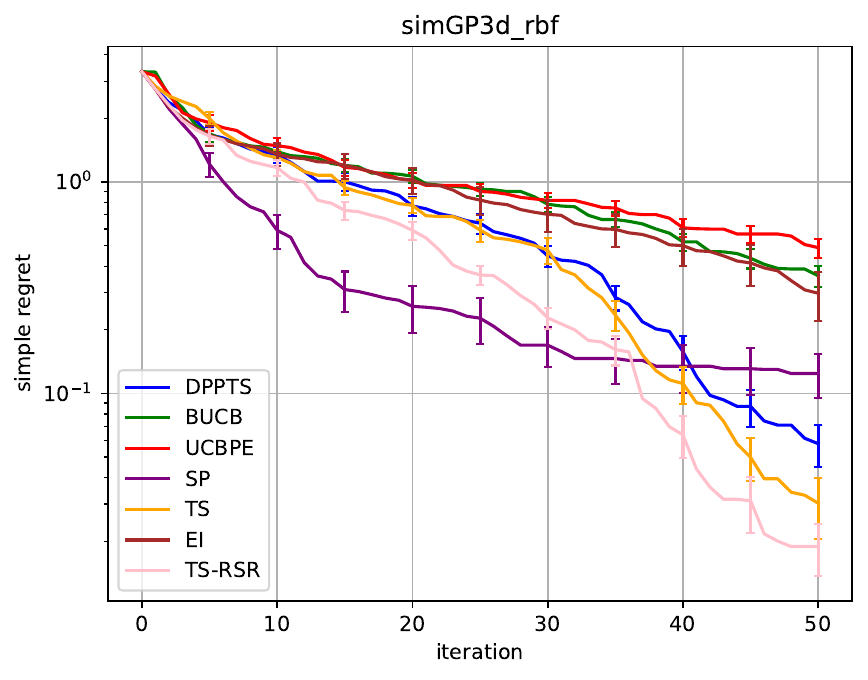}\hfill
  \caption{Simple regret for synthetic functions with known prior. Each curve is the average of 10 runs. The error bars represent $\pm$ 1 standard error. }
  \label{fig:known_prior_regret_traj}
\end{figure*}

To better understand the performance of our algorithm, we first evaluated its performance on functions sampled from a known GP prior. To this end, we 1) sampled 10 random 2D functions from a RBF prior with lengthscale = 0.25, defined on the domain $[-5,5]^2$, and sampled 10 random 3D functions from a GP prior, with a Gaussian RBF kernel that has lengthscale 0.15, defined on the domain $[0,1]^3$. For the 2D function, for each of the ten functions, we repeat each algorithm for ten runs, yielding a total of 100 trials for each algorithm. For the 3D function, for each of the ten functions, we repeat each algorithm for five runs, yielding a total of 50 trials for each algorithm. Before each run, each algorithm has access to 15 random samples, which is identical across all the algorithms. We note that it is nontrivial to compute the standard deviation across the different functions, but in this case, we compute the means and standard deviations in Table \ref{tab:results_gp_prior} by treating each trial as coming from the same function. We see in Table \ref{tab:results_gp_prior} that TS-RSR outperforms its peers in both the 2D and 3D case with known GP prior. The trajectories of simple regret are shown in Figure \ref{fig:known_prior_regret_traj}. We note that considering the total number of available function evaluations ($400$ in the 2D case and $250$ in the 3D case) , both settings are rather difficult considering their domain size and GP prior lengthscale, and given the large number of trials, these serve as representative demonstrations of the superior efficacy and consistency of the proposed TS-RSR algorithm.

\begin{table}[htbp]
\centering
\footnotesize
\caption{Simple regret at last iteration (2D/3D synthetic functions)}
\label{tab:results_gp_prior}
\begin{tabular}{|c|c|c|}
\hline
 & GP-RBF-prior-2D & GP-RBF-prior-3D \\
 \hline
(batch size $m$) & $m = 20$ &$m=5$ \\
\hline
(iterations $T$) & $T = 20$ &$T=50$ \\

\hline
(units for regret)  & $10^{-2}$ &$10^{-2}$ \\ 
\hline
DPPTS & 6.1 ($\pm$11.4) [R: 3] & 5.8 ($\pm$9.1) [R: 3] \\
BUCB & 5.1 ($\pm$11.3) [R: 2] & 36.0 ($\pm$29.5) [R: 6] \\
UCBPE & 11.8 ($\pm$16.8) [R: 5] & 48.9 ($\pm$35.5) [R: 7] \\
SP &  12.7 ($\pm$18.6) [R: 6] & 12.4 ($\pm$20.6) [R: 4] \\
TS & 8.9 ($\pm$14.6) [R: 4] & 3.0 ($\pm$6.8) [R: 2] \\
qEI & 29.8 ($\pm$26.1) [R: 7] & 29.8 ($\pm$54.2) [R: 5] \\
\textbf{TS-RSR} & \textbf{3.8($\pm$10.0) [R: 1]} & \textbf{1.9($\pm$3.6) [R: 1]} \\
\hline
\end{tabular}
\end{table}
\normalsize

\subsection{Synthetic test functions}

\subsubsection{2D/3D functions}

\begin{figure*}[h]
  \centering
  \includegraphics[width=.24\textwidth]{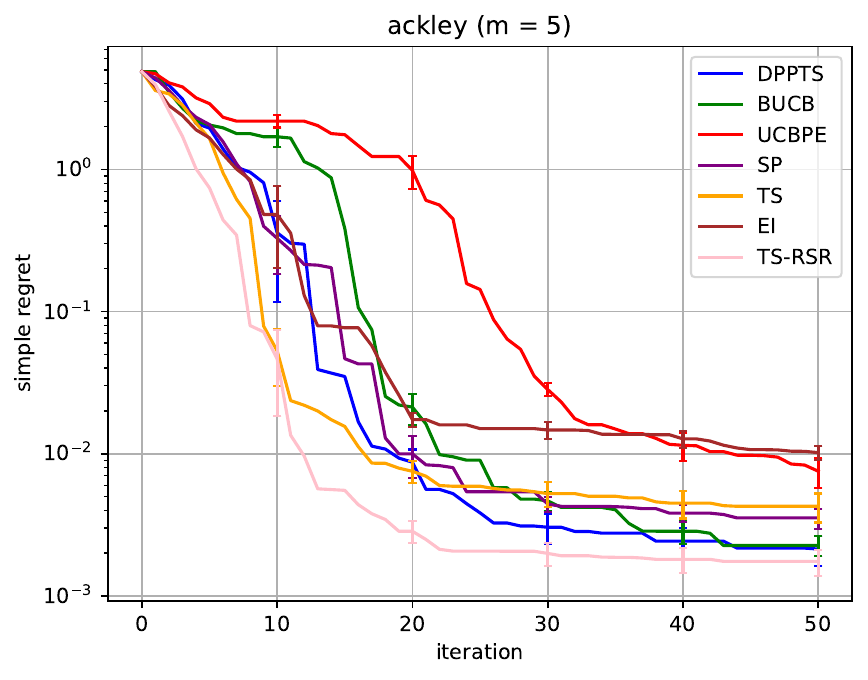}\hfill
\includegraphics[width=.24\textwidth]{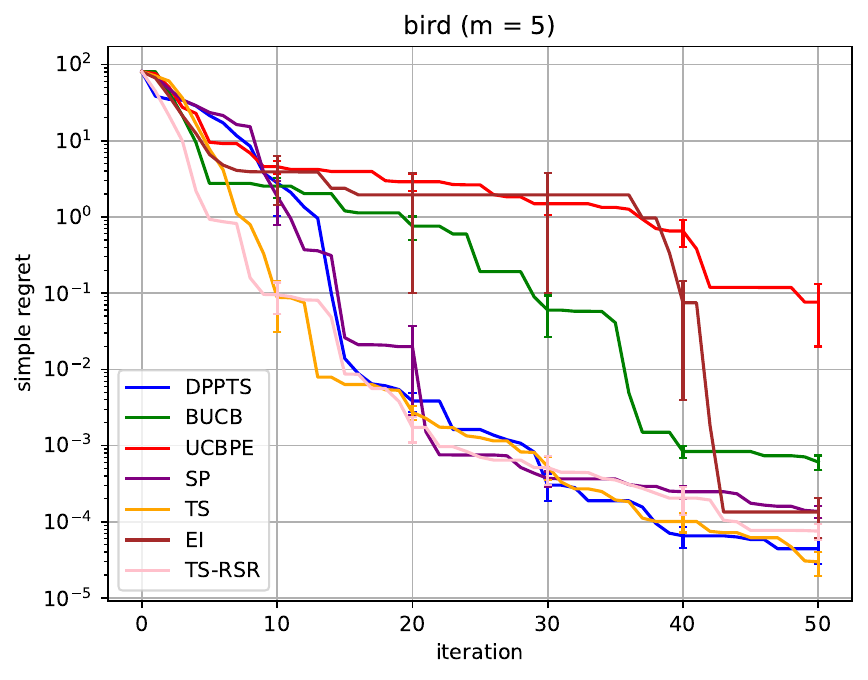}\hfill
\includegraphics[width=.24\textwidth]{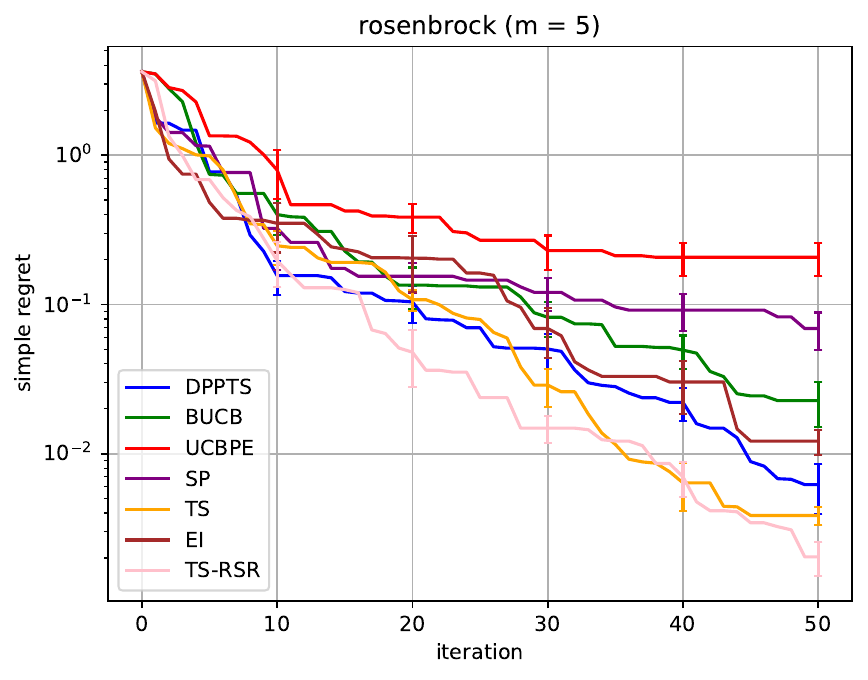}
\includegraphics[width=.24\textwidth]{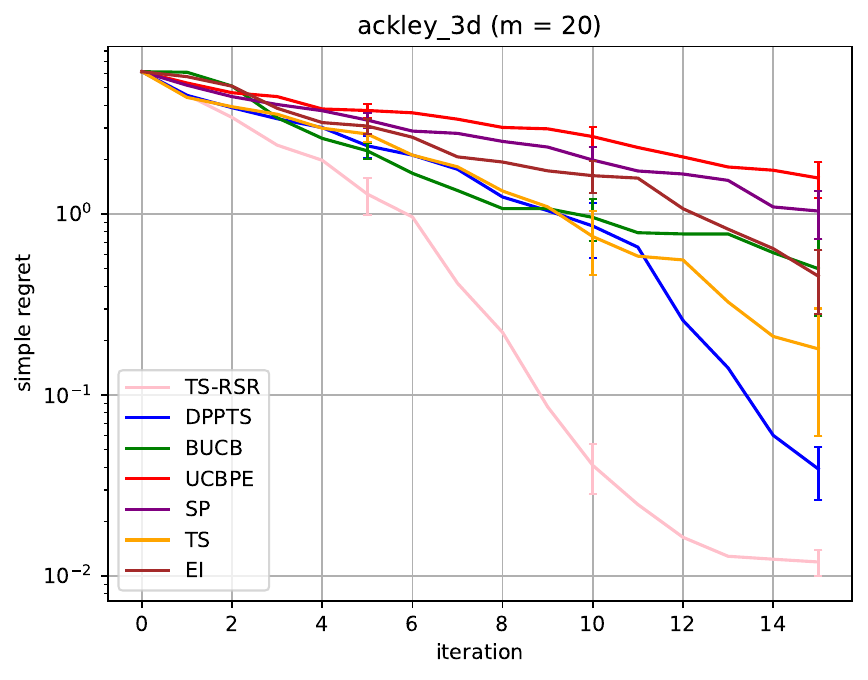}
  \caption{Simple regret for 2D/3D synthetic functions. Each curve is the average of 10 runs. The error bars represent $\pm$ 1 standard error.}
  \label{fig:2d_3d_regret_traj}
\end{figure*}

For the synthetic test functions, we chose from a range of challenging nonconvex test functions, across varying dimensions. In 2D, we have Ackley, Bird, and Rosenbrock. In 3D, we have the 3D version of Ackley. Our results are summarized in Table \ref{tab:results_2d_3d}. As we can see, our algorithm outperforms all the other algorithms for all the test functions here except the Bird, where it performs only slightly worse than TS and DPPTS. The plots of the averaged simple regret for the different algorithms on these test functions can be found in Figure \ref{fig:2d_3d_regret_traj}.

\begin{table}[htbp]
\centering
\scriptsize
\caption{Simple regret at last iteration (2D/3D synthetic functions)}
\label{tab:results_2d_3d}
\begin{tabular}{|c@{\hskip 0pt}|c@{\hskip 0pt}|c@{\hskip 0pt}|c@{\hskip 0pt}|c@{\hskip 0pt}|}
\hline
 & Ackley-2D & Rosenbrock-2D & Bird-2D & Ackley-3d \\
\hline
(batch size $m$) & $m=5$ & $m=5$ & $m=5$ & $m=20$ \\
\hline
(iterations $T$) & $T=50$ & $T=50$ & $T=50$ & $T=15$ \\
\hline
(units for regret) & $10^{-3}$ & $10^{-3}$ & $10^{-4}$ & $10^{-2}$ \\ 
\hline
DPPTS & 2.2($\pm$1.6) [R: 2] & 6.2($\pm$7.2) [R: 3] & 0.4($\pm$1.0) [R: 2] & 3.9($\pm$4.0) [R: 2]  \\
BUCB & 2.3($\pm$1.1) [R: 3] & 22.7($\pm$24.1) [R: 5] & 7.1($\pm$7.2) [R: 6] & 50.1($\pm$71.7) [R: 5]  \\
UCBPE & 8.3($\pm$5.4) [R: 6] & 207.1($\pm$165.9) [R: 7] & 763.3($\pm$1782.0) [R: 7] & 158.7($\pm$113.3) [R: 7]  \\
SP & 3.5($\pm$1.8) [R: 4] & 69.0($\pm$61.1) [R: 6] & 1.4($\pm$1.0) [R: 5] & 104.2($\pm$97.3) [R: 6]  \\
TS & 4.3($\pm$3.1) [R: 5] & 3.9($\pm$1.7) [R: 2] & \textbf{0.3($\pm$0.0) [R: 1]} & 18.0($\pm$38.3) [R: 3]  \\
qEI & 10.4($\pm$3.5) [R: 7] & 12.1($\pm$7.3) [R: 4] & 1.4($\pm$2.0) [R: 4] & 45.6($\pm$55.6) [R: 4] \\
\textbf{TS-RSR} & \textbf{1.7($\pm$ 1.1) [R: 1]} & \textbf{2.0($\pm$ 1.6) [R: 1]} & 0.7($\pm$1.0) [R: 3] & \textbf{1.2($\pm$0.6) [R: 1]} \\
\hline
\end{tabular}
\end{table}
\normalsize

\subsubsection{Higher-dimensional test functions}

\begin{figure*}[h]
  \centering
  \includegraphics[width=.32\textwidth]{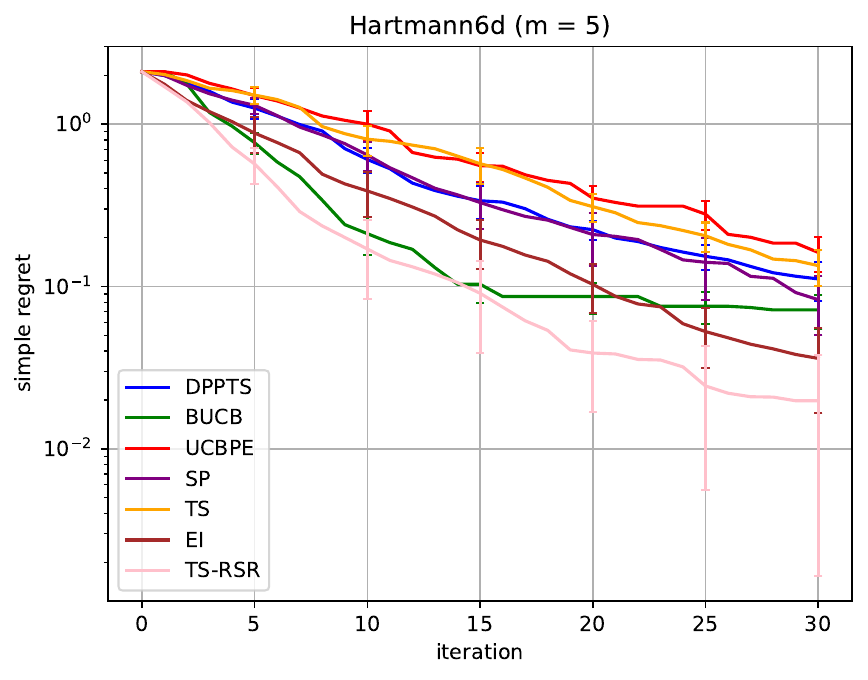}\hfill
\includegraphics[width=.32\textwidth]{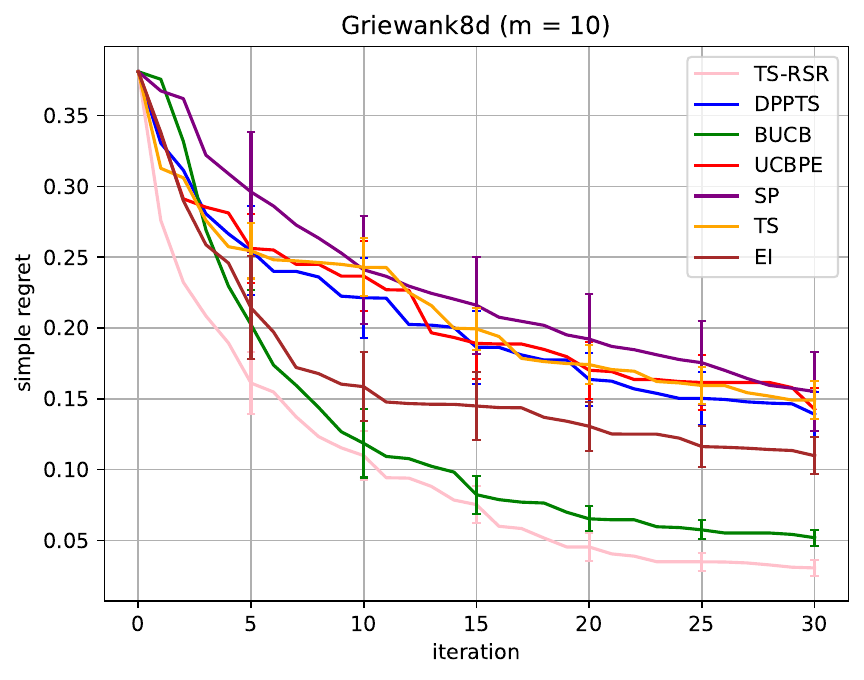}\hfill
\includegraphics[width=.32\textwidth]{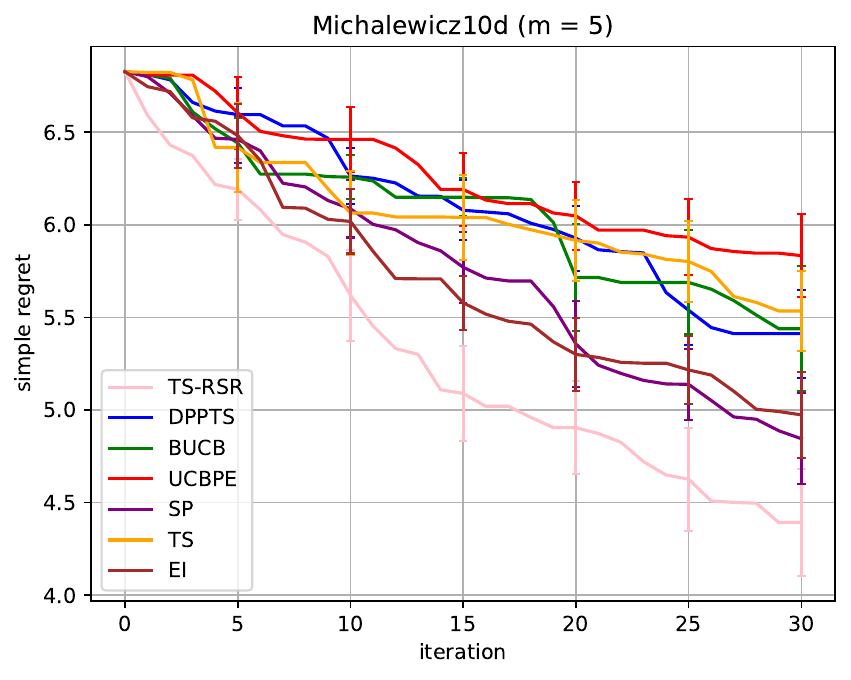}
  \caption{Simple regret for higher-dimensional synthetic functions. Each curve is the average of 10 runs. The error bars represent $\pm$ 1 standard error. }
  \label{fig:highD_regret_traj}
\end{figure*}

We also tested on the following higher dimensional test functions: Hartmann (6D), Griewank (8D), and Michalewicz (10D), which are well-known nonconvex test functions with many local optima. Our results are summarized in Table \ref{tab:results_highD}, and the simple regret curves can be found in Figure \ref{fig:highD_regret_traj}. Again, our proposed algorithm consistently outperforms its competitors.

\begin{table}[h]
\centering
\footnotesize
\caption{Simple regret at last iteration (higher-dimensional)}
\label{tab:results_highD}
\begin{tabular}{|c|c|c|c|}
\hline
 & Hartmann-6D & Griewank-8D & Michaelwicz-10D  \\
\hline
(batch size $m$) & $m=5$ & $m=10$ & $m=5$  \\
\hline
(iterations $T$) & $T=30$ & $T=30$ & $T=30$  \\
\hline
(units for regret) & $10^{-2}$ & $10^{-2}$ & $10^{0}$ \\
\hline
DPPTS & 11.1 ($\pm$ 9.4) [R: 7] & 14.0 ($\pm$ 5.0) [R: 4] & 5.4 ($\pm$ 0.8) [R: 4]  \\
BUCB & 4.78 ($\pm$ 4.9) [R: 4] & 5.2 ($\pm$ 1.8) [R: 2] & 5.4 ($\pm$ 1.1) [R: 4]  \\
UCBPE & 8.3 ($\pm$ 6.5) [R: 6] & 14.3 ($\pm$ 4.8) [R: 6] & 5.8 ($\pm$ 0.7) [R: 7]  \\
SP & 4.0 ($\pm$ 8.0) [R: 3] & 15.5 ($\pm$ 8.9) [R: 7] & 4.8 ($\pm$ 0.8) [R: 2] \\
TS & 5.9 ($\pm$ 9.5) [R: 5] & 14.9 ($\pm$ 4.3) [R: 5] & 5.5 ($\pm$ 0.7) [R: 6]  \\
qEI & 1.9 ($\pm$ 5.5) [R: 2] & 11.0 ($\pm$ 4.1) [R: 3] & 5.0 ($\pm$ 0.7) [R: 3]  \\
TS-RSR & \textbf{1.6 ($\pm$ 4.7) [R: 1]} & \textbf{3.1 ($\pm$ 1.7) [R: 1]} & \textbf{4.4 ($\pm$ 0.7) [R: 1]} \\
\hline
\end{tabular}
\end{table}
\normalsize

\subsection{Real-world test functions}

To better evaluate our algorithm, we also experimented on three realistic real world test functions. 

\begin{figure*}[h]
  \centering
  \includegraphics[width=.32\textwidth]{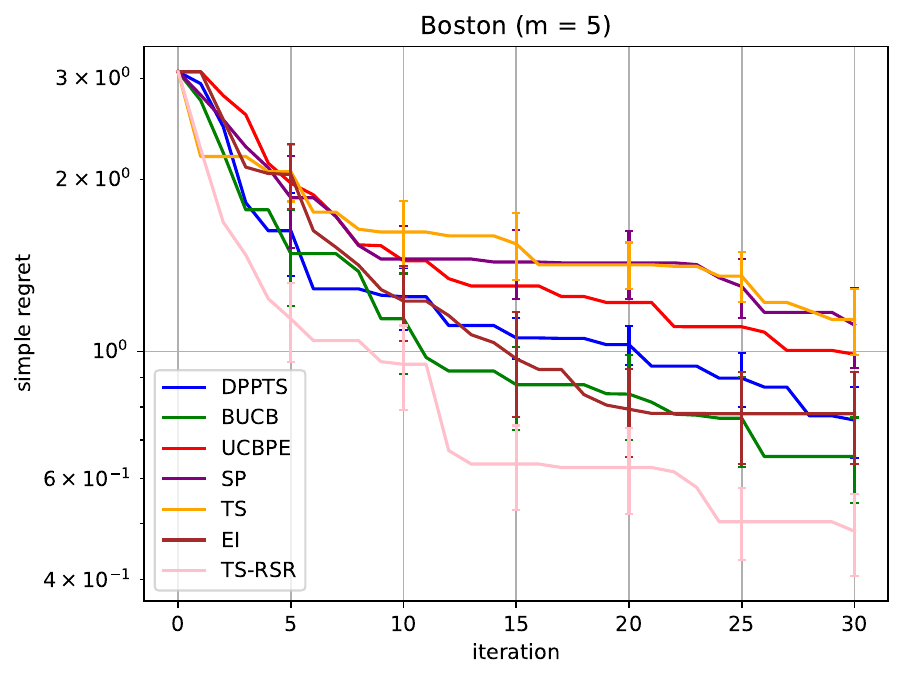}\hfill
\includegraphics[width=.32\textwidth]{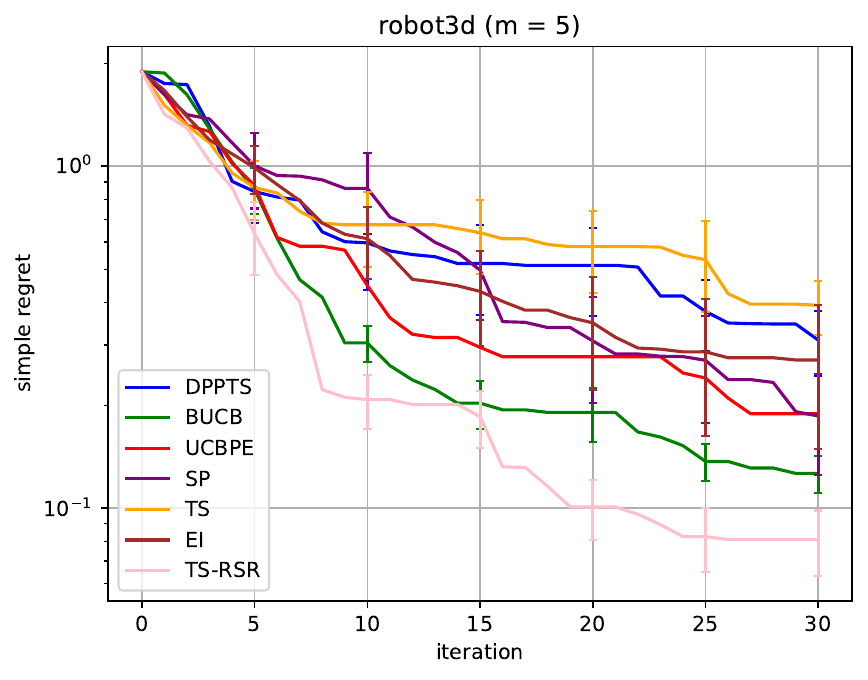}\hfill
\includegraphics[width=.32\textwidth]{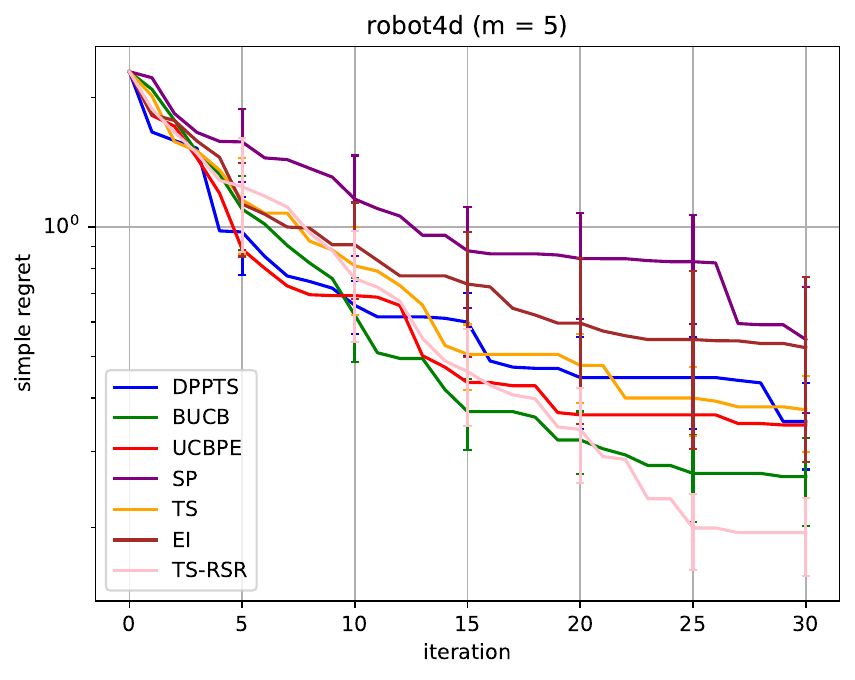}\hfill
  \caption{Average simple regret for Boston housing, robot pushing 3D and robot pushing 4D problems. Each curve is the average over ten runs. The error bars represent $\pm$ 1 standard error.}
  \label{fig:real_traj}
\end{figure*}

First, we have a 4D hyperparameter tuning task for the hyperparameters of the RMSProp optimizer in a 1-hidden layer NN regression task for the Boston housing dataset. Here, the 4 parameters we tune are 1) the number of nodes in the hidden layer (between 1 and 100), 2) the learning rate of the RMSProp optimizer (between $0.001$ and $0.1$), 3) the weight decay of the optimizer (between $0$ and $0.5$), 4) the momentum parameter of the optimizer (between $0$ and $0.5$). The experiment is repeated 10 times, and the neural network’s weight initialization and all other
parameters are set to be the same to ensure a fair comparison. The dataset was randomly split into train/validation sets. We initialize the observation set
to have 15 random function evaluations which were set to be the same across all the methods. The performances of the different algorithms in terms of the simple
regret\footnote{Since a grid search is infeasible over the 4-dimensional search space, to compute the average regret, we take the best validation loss found across all the runs of all the algorithms as our proxy for the best possible loss.} at the last iteration for the regression L2-loss on the validation set of the
Boston housing dataset is shown in Table \ref{tab:results_real_world}. As we can see, TS-RSR outperforms all its competitors, improving on its closest competitor (BUCB) by 25.7$\%$. The trajectories of the average simple regret is shown in Figure \ref{fig:real_traj}.

\begin{table}[htbp]
\centering
\footnotesize
\caption{Simple regret at last iteration (real-world test functions)}
\label{tab:results_real_world}
\begin{tabular}{|c@{\hskip 1pt}|c@{\hskip 1pt}|c@{\hskip 1pt}|c@{\hskip 1pt}|}
\hline
 & (Boston housing) NN regression & Robot pushing (3D) & Robot pushing (4D) \\
\hline
(batch size $m$) & $m=5$ & $m=5$ & $ m = 5$ \\
\hline
(iterations $T$) & $T=30$ & $T=30$ & $T = 30$ \\
\hline
(units for regret) & $10^{-1}$ & $10^{-2}$ & $10^{-1}$ \\
\hline
DPPTS &   7.6 ($\pm$ 3.4) [R: 3] & 31.0 ($\pm$ 20.1) [R: 6] & 3.5($\pm2.5$) [R:4] \\
BUCB & 6.6 ($\pm$ 3.5) [R: 2] & 12.6 ($\pm$ 5.0) [R: 2] & $2.6 (\pm 1.9)$ [R:2] \\
UCBPE &  9.9 ($\pm$ 4.3) [R: 5] & 18.9 ($\pm$ 6.3) [R: 4] & $3.5 (\pm 2.3)$ [R:3] \\
SP &  11.1 ($\pm$ 5.6) [R: 6] & 18.6 ($\pm$ 19.2) [R: 3] & 5.5 ($\pm(5.6)$)[R:7] \\
TS &  11.4 ($\pm$ 4.8) [R: 7] & 39.2 ($\pm$ 22.3) [R: 7] & 3.8 ($\pm$ 2.4)[R:5] \\
qEI &  7.8 ($\pm$ 4.5) [R: 4] & 27.1 ($\pm$ 38.7) [R: 5] & $5.2$($\pm7.6$)[R:6] \\
TS-RSR & \textbf{4.9 ($\pm$ 2.5) [R: 1]} & \textbf{8.1 ($\pm$ 5.5) [R: 1]} & \textbf{1.9($\pm$ 1.3) [R:1]} \\
\hline
\end{tabular}
\end{table}
\normalsize

Next, we experimented on the active learning for robot pushing setup from [4]. This consists of conducting active policy search on the task of selecting a pushing action of an object towards a designed goal location. There are two variants to the problem with one being 3D, and another being 4D. For the 3D function, the input includes the robot location $(r_x,r_y)$ and the pushing duration $t_r$; for the 4D, the input also includes specifying the initial angle the robot faces. In this experiment, we also have ten repetitions for both the two functions, where each repetition represents a different goal. The simple regret performances at the last iteration can be found in Table \ref{tab:results_real_world}, where we again we see that TS-RSR significantly outperforms its peers, improving on its closests competitor (BUCB in both cases) by 35.7$\%$ in the 3D case and $25.7\%$ in the 4D case respectively. The trajectories of the average simple regret is shown in Figure \ref{fig:real_traj}. We provide more details of our experimental setup in Appendix \ref{appendix:simulation_details}.

 \vspace{-0.5em}
\section{Conclusion}\label{sec:conclusion}
 \vspace{-0.25em}

In this paper, we introduced a new algorithm, $\alg$, for the problem of batch BO. We provide strong theoretical guarantees for our algorithm via a novel analysis, which may be of independent interest to researchers interested in studying IDS methods for BO. Moreover, we confirm the efficacy of our algorithm on a range of simulation problems, where we attain strong, state-of-the-art performance. We believe that our algorithm can serve as a new benchmark in batch BO, and as a buiding block for more effective batch BO in practical applications.
\appendix
 \vspace{-0.5em}
\section{Useful results for Theorem \ref{theorem:main_result}}\label{appendix:main_result}
 \vspace{-0.25em}

\subsection{Information-theory result for Lemma \ref{lemma:sum_sigma_informational bound}}\label{appendix:information_theory}

We have the following result (Lemma 5.3 in \cite{srinivas2009gaussian}), which states that the information gain for any set of selected points can be expressed in terms of posterior variances. This result is useful in the proof of Lemma \ref{lemma:sum_sigma_informational bound}, which is in turn key to the bound for the RSR which we use to show Theorem \ref{theorem:main_result}.
\begin{lemma}
\label{lemma:info_gain_sigma_rship}
For any positive integer $T$, denoting $\bm{f}_{[T]}$ as $\{f(x_{i})\}_{i=1}^T$ and $\bm{y}_{[T]}$ as $\{y_{i}\}_{i=1}^T $, where $y_i = f(x_i) + \epsilon_i$ and $\epsilon_i \sim N(0,\sigma_n^2)$, we have
\footnotesize
\begin{align*}
    I(\bm{y}_{[T]}; \bm{f}_{[T]}) = \frac{1}{2}\sum_{i=1}^T \log\left(1 +\sigma_n^{-2}\sigma_{i-1}^2(x_i) \right) = \sum_{i=1}^T I(f; y_i | \bm{y}_{[i-1]}).
\end{align*}
\normalsize
\end{lemma}
\begin{proof}
    For completeness, we provide the result below here. Using standard information theory identities, we have
    \footnotesize
    \begin{align*}
         I(\bm{y}_{[T]}; \bm{f}_{[T]}) 
        \labelrel{=}{eq:MI_H_identity} & \ H(\bm{y}_{[T]}) - H(\bm{y}_{[T]} \mid \bm{f}_{[T]}) \\
       \labelrel{=}{eq:entropy_decomp} & \ \left(H(y_T \mid \bm{y}_{[T-1]}) + H(\bm{y}_{[T-1]})\right) - \left(H(y_T \mid \bm{f}_{[T]}, \bm{y}_{[T-1]}) + H(\bm{y}_{[T-1]} \mid \bm{f}_{[T]})\right) \\
        \labelrel{=}{eq:cond_entropy_simplifies} & \ \left(H(y_T \mid \bm{y}_{[T-1]}) + H(\bm{y}_{[T-1]})\right) - \left(H(y_T \mid f(x_T)) + H(\bm{y}_{[T-1]} \mid \bm{f}_{[T-1]})\right) \\
        = & \ \left(H(y_T \mid \bm{y}_{[T-1]}) - H(y_T \mid f(X_T))\right) + \left(H(\bm{y}_{[T-1]}) - H(\bm{y}_{[T-1]} \mid \bm{f}_{[T-1]})\right) \\
        \labelrel{=}{eq:diff_entropy_identity} & \ \left( \frac{1}{2}\left(\log(\sigma_{T-1}^2(x_{T}) + \sigma_n^2)  - \log(\sigma_n^2)) \right)\right)  + I(\bm{y}_{[T-1]}; \bm{f}_{[T-1]}) \\
        = & \ \frac{1}{2}\log(1 + \sigma_n^{-2} \sigma_{T-1}^2(x_{T}))    + I(\bm{y}_{[T-1]}; \bm{f}_{[T-1]}) \\
        = & \ \dots \\
        \labelrel{=}{eq:iterate} & \ \frac{1}{2} \sum_{i=1}^T \log \left(1 + \sigma_n^{-2} \sigma_{i-1}^2(x_i) \right) = \sum_{i=1}^T \left(H(y_i \mid \bm{y}_{[i-1]}) - H(y_i \mid f(x_i))\right) \\
        = & \  \sum_{i=1}^T I(y_i; f(x_i) | \bm{y}_{[i-1]})  = \sum_{i=1}^T I(y_i; f | \bm{y}_{[i-1]}) 
    \end{align*}
    \normalsize
\end{proof}

In the derivations above, (\ref{eq:MI_H_identity}) follows from the identity $I(Y;X) = H(Y) - H(Y\mid X)$ which holds for any random variables $X$ and $Y$, where $I(Y;X)$ denotes the mutual information between the variables $Y$ and $X$, $H(\cdot)$ denotes (differential) entropy. Meanwhile, (\ref{eq:entropy_decomp}) holds since for any random variables $X_1$ and $X_2$, we have the identity $H(X_1, X_2) = H(X_1) + H(X_2 \mid X_1)$; (\ref{eq:cond_entropy_simplifies}) holds since $y_T$ is independent of $\bm{y}_{[T-1]}$ and $\bm{f}_{[T-1]}$ conditional on $f(x_T)$, and similarly $\bm{y}_{T-1}$ is independent of $f(x_T)$ conditional on $\bm{f}_{[T-1]}$. The equality (\ref{eq:diff_entropy_identity}) follows from the fact that $y_T \mid \bm{y}_{[T-1]}$ follows a Gaussian distribution with variance $\sigma_n^2 + \sigma_{T-1}^2(x_T)$ and $y_T \mid f(x_T)$ follows a Gaussian distribution with variance $\sigma_n^2$, and the fact that the differential entropy of a Gaussian with any variance $\sigma^2$ is equal to $\frac{1}{2}(\log(2\pi e \sigma^2))$. The equation (\ref{eq:iterate}) follows from iterating the derivations and applying them iteratively on $I(\bm{y}_{[T-1]}; \bm{f}_{[T-1]})$, $\dots$, $I(\bm{y}_{[1]}; \bm{f}_{[1]})$. The final equality simply states that that the mutual information can then be written as a sum of conditional mutual information gains, which will be useful in streamlining some later analysis.

\subsection{Bounding $\xpt{\tilde{R}_{T,m}^2}$, which is useful to prove Theorem \ref{theorem:main_result}}
\label{appendix:bound_xpt_tilde_R_sq}

We next focus on bounding $\xpt{\tilde{R}_{T,m}^2}$, where we recall that $\tilde{R}_{T,m} := \sum_{t=0}^{T-1} \sum_{i=1}^m \tilde{f}_{t,i}^* - f(x_{t,i}^\alg)$. This bound is important in the proof of Theorem \ref{theorem:main_result}. To achieve this, we first introduce the following technical result, which gives us a helpful bound relating to a (discrete) Gaussian Process with $D$ elements.

\begin{lemma}
    \label{lemma:variance_sup_GP_bdd}
    Consider a $D$-dimensional Gaussian, $\bm{Y} \sim N(0, \bm{\Sigma})$, where $\bm{\Sigma} \succ \bm{0}_{D \times D}$. Suppose that $D \geq 2$, and that for each $j \in [D]$, we have $\sigma_j^2 \leq 1$, where $\sigma_j^2 := \bm{\Sigma}_{j,j}$. Then,
$\xpt{\max_{j \in [D]}(\bm{Y}_j)^2} \leq 6\log D.$
\end{lemma}
\begin{proof}
    First, observe that for each $i \in [D]$, applying the standard formula for the moment-generating-function (MGF) of a chi-squared random variable, we have $\xpt{\exp(\lambda \bm{Y}_i^2)} = \frac{1}{\sqrt{1 - 2\lambda \sigma_i^2}}$ whenever $\lambda \leq \frac{1}{2\sigma_i^2}$. Since $\sigma_i^2 \leq 1$, for any $\lambda  < \frac{1}{2}$, we have $ \xpt{\exp(\lambda \bm{Y}_i^2)} = \frac{1}{\sqrt{1 - 2\lambda \sigma_i^2}} \leq \frac{1}{\sqrt{1 - 2\lambda}}$. Now, for any $\lambda < 1/2$, observe that
\footnotesize
    \begin{align*}
 \exp\left(\lambda \xpt{\max_{i \in [D]} (\bm{Y}_i)^2}\right) 
        \labelrel{\leq }{eq:Jensen's_sup_var} & \ \xpt{\exp(\lambda \max_{i \in [D]}(\bm{Y}_i)^2)} = \xpt{\max_{i \in [D]}\exp(\lambda (\bm{Y}_i)^2)} \\
        \leq & \   \xpt{\sum_{i=1}^D \exp(\lambda (\bm{Y}_i)^2)} =  \sum_{i=1}^D \xpt{\exp(\lambda (\bm{Y}_i)^2)} \leq \ \frac{D}{\sqrt{1 - 2\lambda}} 
    \end{align*}
    \normalsize
    Above, to derive (\ref{eq:Jensen's_sup_var}) we used Jensen's equality. Taking log on both sides and dividing by $\lambda$, we find that
    \footnotesize
    \begin{align*}
        \xpt{\max_{i \in [D]} \bm{Y_i}^2} \leq \frac{\log(D) - \frac{1}{2}\log(1 - 2\lambda)}{\lambda}.
    \end{align*}
    \normalsize
    Setting $\lambda = \frac{1}{4}$ (which is less than $\frac{1}{2}$), we then find that
    \footnotesize
    \begin{align*}
        \xpt{\max_{i \in [D]}\bm{Y_i}^2} \leq 4 \left(\log(D) + \frac{1}{2}\log(2)\right) \leq 6 \log D,
    \end{align*}
    \normalsize
    where the final inequality uses the assumption that $D \geq 2$.
\end{proof}

Equipped with the above technical result, we are now ready to bound $\xpt{\tilde{R}_{T,m}^2}$.
\begin{lemma}
    \label{lemma:R_tilde_^2_bdd}
    Suppose that $\abs*{\mathcal{X}} \geq 2$, and that $k(x,x) \leq 1$ for any $x \in \mathcal{X}$. Then, for any $t \in [T]$ and $i \in [m]$, we have
    \footnotesize
    \begin{align*}
        \mathbb{E}[(\tilde{f}_{t,i}^* - f(x_{t,i}^\alg))^2] \leq 24 \log \mathcal{X},
    \end{align*}
    \normalsize
    which implies then that
    \footnotesize
    \begin{align*}
        \xpt{\tilde{R}_{T,m}^2} =  \xpt{\left(\sum_{t=0}^{T-1} \sum_{i=1}^m \tilde{f}_{t,i}^* - f(x_{t,i}^\alg)\right)^2} 
        \leq & \  (Tm)^2 \sum_{t=0}^{T-1} \sum_{i=1}^m  \mathbb{E}[(\tilde{f}_{t,i}^* - f(x_{t,i}^\alg))^2] \\ 
        \leq & \  24 \log \abs*{\mathcal{X}} (Tm)^3
    \end{align*}
    \normalsize
    
\end{lemma}
\begin{proof}
    We observe that for any $t \in [T]$ and $i \in [m]$,
    \footnotesize
    \begin{align}
    & \ \mathbb{E}[(\tilde{f}_{t,i}^* - f(x_{t,i}^\alg))^2] \leq   2 \mathbb{E}[(\tilde{f}_{t,i}^*)^2]  + 2 \mathbb{E}[(f(x_{t,i}^\alg))^2] \nonumber \\
    \labelrel{=}{eq:tower} & \ 2 \xpt{\xpt{(\tilde{f}_{t,i}^*)^2 \mid \mathcal{F}_t}} + 2 \xpt{(f(x_{t,i}^\alg))^2} 
    \labelrel{=}{eq:ts_dist_same}  2 \xpt{\xpt{(f^*)^2 \mid \mathcal{F}_t}} + 2 \xpt{(f(x_{t,i}^\alg))^2} \nonumber \\
    \labelrel{=}{eq:tower_2} & \ 2 \xpt{(f^*)^2} +  2 \xpt{(f(x_{t,i}^\alg))^2} \labelrel{\leq}{eq:apply_var_sup_GP_bdd}  24 \log \abs*{\mathcal{X}} \label{eq:xpt_tilde_diff_sq_bdd}
    \end{align}
    \normalsize
    Above, we used the tower property of conditional expectation in (\eqref{eq:tower}), and in (\eqref{eq:ts_dist_same}), we used the fact that $f^* \mid \mathcal{F}_t$ has the same distribution as $\tilde{f}_{t,i}^* \mid \mathcal{F}_t$, which implies also that $(f^*)^2 \mid \mathcal{F}_t$ has the same distribution as $(\tilde{f}_{t,i}^*)^2 \mid \mathcal{F}_t$. The equation (\ref{eq:tower_2}) follows again from the tower property, while to derive (\ref{eq:apply_var_sup_GP_bdd}), we used Lemma \ref{lemma:variance_sup_GP_bdd}, which we just proved. The final bound on $\xpt{\tilde{R}_{T,m}^2}$ then follows from Cauchy-Schwarz and applying \eqref{eq:xpt_tilde_diff_sq_bdd}.
\end{proof}

\section{Bounds on $\rho_m$ through an initialization strategy and bounds on information gain term}

\subsection{Bounding the $\rho_m$ term through an initialization strategy}\label{appendix:rho_m_bound}

Recall that $\rho_m := \max_{x \in \mathcal{X}, \tau, \tilde{A}_m \subset \mathcal{X}, \abs*{\tilde{A}_m} \leq M} \frac{\sigma_{\tau}(x)}{\sigma_{\tau}(x \mid \tilde{A}_m)}$, which denotes the maximal decrease in posterior uncertainty resulting from conditioning on any additional set of samples $\tilde{A}_m$ up to $m$. 

In order for the regret bound in Theorem \ref{theorem:main_result} to scale sublinearly in $m$, we need $\rho_m = o(\sqrt{m})$. We will show that via a two-step procedure where we first initialize following a maximal variance strategy and then run the algorithm, the term $\rho_m$ can in fact be made to be $\tilde{O}(1)$. We note that our results here are largely a restatement of their counterparts in \cite{desautels2014parallelizing}, and are provided here for completeness and for the reader's convenience. 

First, we will bound $\rho_m$ in terms of a mutual information quantity, which we will later show can be bound tractably when an appropriate initialization strategy is used.

\begin{lemma}
\label{lemma:rho_m_bdd_by_conditional_MIG}
Suppose we have an initialization set $D_{T_{init}} \subset \mathcal{X}$ of size $T_{init}$, which we sample before running the algorithm. Then, defining
\footnotesize
\begin{align*}
    \rho_m(D_{T_{init}}) = \max_{x \in \mathcal{X}, \tau, \tilde{A}_m \subset \mathcal{X}, \abs*{\tilde{A}_m} \leq M} \frac{\sigma_{\tau}(x \mid D_{T_{init}})}{\sigma_{\tau}(x \mid \tilde{A}_m, D_{T_{init}})},
\end{align*}
\normalsize
by defining $C_{m}(D_{T_{init}}) := \max_{\tilde{A}_m \subset \mathcal{X}, \abs*{\tilde{A}_m} \leq m} I\left(f; \bm{y}_{\tilde{A}_m} \mid \bm{y}_{D_{T_{init}}}\right)$, we have
\footnotesize
\begin{align*}
    \rho_m(D_{T_{init}}) \leq \exp\left(\max_{\tilde{A}_m \subset \mathcal{X}, \abs*{\tilde{A}_m} \leq m} I\left(f; \bm{y}_{\tilde{A}_m} \mid \bm{y}_{D_{T_{init}}}\right)\right) := \exp\left(C_{m}(D_{T_{init}}) \right),
\end{align*}
\normalsize
\end{lemma}

\begin{proof}
    The proof follows from a straightforward derivation. We note that for any $x \in \mathcal{X}$, and any positive integer $t$ and $\tilde{A}_m \subset \mathcal{X}$ where $\abs*{\tilde{A}_m} \leq m$, 
    \footnotesize
    \begin{align*}
        I(f(x); \bm{y}_{\tilde{A}_m} \mid \bm{y}_{[t]}, \bm{y}_{D_{T_{init}}}) = & \ H(f(x) \mid \bm{y}_{[t]}, \bm{y}_{D_{T_{init}}}) - H(f(x) \mid \bm{y}_{[t]}, \bm{y}_{\tilde{A}_m} ,\bm{y}_{D_{T_{init}}}) \\
        = & \ \frac{1}{2}\log\left(\frac{\sigma_t^2(x \mid D_{T_{init}})}{\sigma_t^2(x \mid \tilde{A}_m, D_{T_{init}})} \right).
    \end{align*}
    \normalsize
Hence, by algebraic manipulation, we see that
\footnotesize
\begin{align*}
    \frac{\sigma_t(x \mid D_{T_{init}})}{\sigma_t(x \mid \tilde{A}_m, D_{T_{init}})} \leq & \ \exp\left(  I(f(x); \bm{y}_{\tilde{A}_m} \mid \bm{y}_{[t]}, \bm{y}_{D_{T_{init}}})\right) \\
    \leq & \ \exp\left(  I(f; \bm{y}_{\tilde{A}_m} \mid \bm{y}_{[t]}, \bm{y}_{D_{T_{init}}})\right) \leq \exp\left(  I(f; \bm{y}_{\tilde{A}_m} \mid \bm{y}_{D_{T_{init}}})\right),
\end{align*}
\normalsize
where the second inequality comes from the fact that $f(x)$ contains strictly less information than $f$, and the final inequality comes from the fact that conditioning always reduces mutual information. The final result then follows from maximizing over all $\tilde{A}_m \subset \mathcal{X}$ with cardinality at most $m$.
\end{proof}

The previous statement shows that we can bound $\rho_m(D_{T_{init}})$ by a term $\exp(C_m(D_{T_{init}}))$, where $C_m(D_{T_{init}}) := \max_{\tilde{A}_m \subset \mathcal{X}, \abs*{\tilde{A}_m} \leq m} I\left(f; \bm{y}_{\tilde{A}_m} \mid \bm{y}_{D_{T_{init}}}\right)$ denotes the maximal mutual information between a set of measurements of size at most $m$ and $f$ conditional on $y_{D_{T_{init}}}$. We now show that by using a initialization strategy where we always sample the point with the maximal posterior variance, the term $C_m(D_{T_{init}})$ can be made be of size $\tilde{O}(1)$, assuming $\gamma_{T_{init}} := \max_{\tilde{A}_{T_{init}}, \abs*{\tilde{A}_{T_{init}}} \leq m} I(f; y_{\tilde{A}_{T_{init}}})$ grows sublinearly with $T_{init}$ and $T_{init}$ is chosen appropriately.

\begin{lemma}
\label{lemma:appendix_initialization_bdd}
    Let $T_{init}$ be the size of the initialization set. For each $j \in [T_{init}]$, let $D_{j}$ denote the first $j$ points in the initialization set. Consider an initialization strategy where for each $i \in [T_{init}]$, we choose $x_i \in \argmax_{x \in \mathcal{X}} \sigma_{i-1}^2(x)$. Then, 
    \footnotesize
    \begin{align*}
        C_m(D_{T_{init}}) := \max_{\tilde{A}_m \subset \mathcal{X}, \abs*{\tilde{A}_m} \leq m} I\left(f; \bm{y}_{\tilde{A}_m} \mid \bm{y}_{D_{T_{init}}}\right) \leq \frac{m}{T_{init}} \gamma_{T_{init}}. 
    \end{align*}
\normalsize
Thus, in combination with Lemma \ref{lemma:rho_m_bdd_by_conditional_MIG}, we have
\footnotesize
\begin{align*}
    \rho_m(D_{T_{init}}) \leq \exp\left(\frac{m \gamma_{T_{init}}}{T_{init}} \right).
\end{align*}
\normalsize
As a corollary, if $\gamma_{T_{init}}$ grows sublinearly with $T_{init}$, then by picking $T_{init}$ such that $T_{init} \geq m \gamma_{T_{init}}$, we have
\footnotesize
\begin{align*}
    \rho_m(D_{T_{init}}) \leq \exp\left(1 \right).
\end{align*}
\normalsize

\end{lemma}
\begin{proof}
    First, we observe that by the choice of initialization, $\sigma_{T_{init}-1}^2(x_{T_{init}}) \leq \sigma_{i-1}^2(x_i)$ for all $i \in [T_{init} - 1]$. In addition, by the initialization choice of maximizing posterior variance, at any round $t > T_{init}$, $\sigma_{t-1}^2(x_t)\leq \sigma_{T_{init}-1}^2(x_{T_{init}})$. Thus, for any $\tilde{A}_m \subset \mathcal{X}$ such that $\abs*{\tilde{A}_m} = m$, by letting $x_{T_{init}+1}, \dots, x_{T_{init} +m}$ denote the $m$ points in $\tilde{A}_m$, we have that 
    \footnotesize
    \begin{align*}
       & \  I(f; \bm{y}_{\tilde{A}_m} \mid \bm{y}_{D_{T_{init}}}) =  \frac{1}{2}\sum_{j=1}^m \log(1 + \sigma_n^{-2} \sigma_{T_{init} + i - 1}^2(x_{T_{init} +i}) \\
        \leq & \ \frac{1}{2}\sum_{j=1}^m \log(1 + \sigma_n^{-2} \sigma_{T_{init}-1}^2(x_{T_{init}})) =  \ m I(f;y_{T_{init}} \mid \bm{y}_{D_{T_{init} - 1}}) \leq  m \frac{I(f;\bm{y}_{D_{T_{init}}})}{T_{init}} \leq \ \frac{m \gamma_{T_{init}}}{T_{init}},
    \end{align*}
    \normalsize
    where the first equality follows from Lemma \ref{lemma:info_gain_sigma_rship}, the first inequality follows from the maximal variance initialization strategy, the second equality follows again from Lemma \ref{lemma:info_gain_sigma_rship} (setting $T$ in Lemma \ref{lemma:info_gain_sigma_rship} to 1). The second inequality follows from the a combination of Lemma \ref{lemma:info_gain_sigma_rship} and the fact that due to the maximal variance initialization strategy,  $\sigma_{T_{init}-1}^2(x_{T_{init}}) \leq \sigma_{i-1}^2(x_i)$ for all $i \in [T_{init} - 1]$. The final inequality follows from the definition of $\gamma_{T_{init}}$ as $\gamma_{T_{init}} = \max_{\tilde{A}_{T_{init}} \subset \mathcal{X}, \abs*{\tilde{A}_{T_{init}}} \leq T_{init}}I(f;\bm{y}_{\tilde{A}_{T_{init}}})$. Combining with Lemma \ref{lemma:rho_m_bdd_by_conditional_MIG}, this completes our proof.
\end{proof}

\subsection{Bounds for the information gain quantity $\gamma_{Tm}$ for different kernels}\label{appendix:gamma_bounds}

We note that following a known result in \cite{srinivas2009gaussian}, $\gamma_{Tm}$ in fact satisfies sublinear growth for three well-known classes of kernels, namely the linear, exponential and Matern kernels.

\begin{lemma}[cf. Theorem 5 in \cite{srinivas2009gaussian}]
    \label{lemma:MIG_kernel_bounds}
For any $\tau > 0$, the maximal information gain $\gamma_{\tau}$ can be bounded as follows for the following kernels.
\footnotesize
    \begin{enumerate}[leftmargin=0.5cm]
        \item (Linear kernel): If $k(x,x')=x^\top x'$, then $$\gamma_{\tau} = O(d \log (\tau)).$$
        \item (Squared exponential kernel): If $k(x,x')=\exp(-\|x-x'\|^2/2)$, then
        $$\gamma_{\tau} = O(\left(\log(\tau)\right)^{d+1}).$$
        \item (Matern kernel with $\nu > 1$): If $k(x,x')= \frac{1}{\Gamma(\nu)2^{\nu-1}} \left( \frac{\sqrt{2\nu}}{d} \|x-x'\| \right)^v K_v\left( \frac{\sqrt{2v}}{d} \|x-x'\| \right)$,
where  $K_v(\cdot)$ is a modified Bessel function, and $\Gamma(\cdot)$ denotes the gamma function, then
        $$\gamma_{\tau} = O( (\tau)^{\frac{d(d+1)}{2\nu + d(d+1)}}\log(\tau))$$
    \end{enumerate}
\normalsize
\end{lemma}

\subsection{Convergence rate of $\alg$ with maximal-variance initialization strategy}\label{appendix:end_to_end}

As discussed in Appendix \ref{appendix:rho_m_bound}, following a technique established in \cite{desautels2014parallelizing}, where we have an exploration phase of length $T_{init}$ where we always sample the point with the highest posterior variance, for sufficiently large $T_{init}$, we may reduce $\rho_m$ to be of size $\tilde{O}(1)$. It is not hard to see that this will come at the expense of a $\tilde{O}(T_{init})$ term in the regret. However, when the horizon $Tm$ is sufficiently large, the $\tilde{O}(T_{init})$ term is dominated by the $\tilde{O}(\sqrt{Tm \gamma_{Tm}})$ term from the second phase. This thus yields the following end-to-end regret bound that grows sublinearly in $Tm$ whenever $\gamma_{Tm}$ grows sublinearly in $Tm$ (as is the case for the linear, squared exponential and Matern kernels), indicating the provable benefit of increasing the batch size $m$. In particular, for the linear and squared exponential kernels, the end-to-end regret scales as $\tilde{O}(\sqrt{Tm})$.

\begin{corollary}
\label{corollary:overall_regret_with_init}
Consider a two-stage algorithm where the initialization stage has $T_{init}$ steps and consider an initialization strategy where for each $i \in [T_{init}]$, we sample $x_i \in \argmax_{x \in \mathcal{X}} \sigma_{i-1}^2(x)$ 
and observe $y_i = f(x_i) + \epsilon_i$. Consider running $\alg$ after the initialization stage for $T$ rounds and a batch size of $m$. Then, slightly abusing notation and (re-)denoting 
$\rho_m := \max_{x \in \mathcal{X}, \tau, \tilde{A}_m \subset \mathcal{X}, \abs*{\tilde{A}_m} \leq m} \frac{\sigma_\tau(x \mid D_{T_{init}})}{\sigma_\tau(x \mid \tilde{A}_m, D_{T_{init}})}$, we have
\footnotesize
\begin{align*}
    \rho_m \leq \exp\left(\frac{m\gamma_{T_{init}}}{T_{init}}\right).
\end{align*}
\normalsize
Thus, whenever $\gamma_{T_{init}}$ grows sublinearly with $T_{init}$, by picking $T_{init}$ sufficiently large such that $\rho_m$ is upper bounded by an absolute constant $C$,  the overall regret (in both phases) satisfies 
\footnotesize
\begin{align*}
    \xpt{R_{T_{init}}} + \xpt{R_{T,m}} = \tilde{O}(T_{init} \sqrt{\log \abs*{\mathcal{X}}}) + \tilde{O}(\sqrt{Tm \gamma_{Tm}\log(\abs*{\mathcal{X}})}),
\end{align*}
\normalsize
where $R_{T_{init}}$ is the cumulative regret from the initialization phase (which lasts for $T_{init}$ steps, and $R_{T,m}$ is the regret incurred in the subsequent $T$ rounds when $\alg$ is used (with a batch size of $m$). Above, $\tilde{O}(\cdot)$ hides polylogarithmic terms in $m$ and $T$.
In particular, for the linear and squared exponential kernel, we can pick $T_{init}$ to be of size $m \operatorname{polylog}(m)$ such that the overall regret satisfies the bound 
\footnotesize
\begin{align*}
    \xpt{R_{T_{init}}} + \xpt{R_{T,m}} = & \  \tilde{O}(m \sqrt{\log \abs*{\mathcal{X}}}) + \tilde{O}(\sqrt{Tm \gamma_{Tm}\log(\abs*{\mathcal{X}})}) \\
    = & \ \begin{cases}
   \tilde{O}(m \sqrt{\log \abs*{\mathcal{X}}}) + \tilde{O}\left(\sqrt{ Tm d \log(Tm) \log(\abs*{\mathcal{X}})}\right) & \mbox{linear kernel} \\
   \tilde{O}(m\sqrt{\log \abs*{\mathcal{X}}}) + \tilde{O}\left(\sqrt{Tm (\log(Tm))^d \log(\abs*{\mathcal{X}})} \right) & \mbox{sq. exp. kernel}.
    \end{cases}
\end{align*}
\normalsize
Meanwhile, for the Matern kernel with parameter $\nu >1$, we can pick $T_{init}$ to be of size $\mathrm{poly}(m)$ such that the overall regret satisfies the bound 
\footnotesize
\begin{align*}
    \xpt{R_{T_{init}}} + \xpt{R_{T,m}} = \tilde{O}(\operatorname{poly}(m)\sqrt{\log \abs*{\mathcal{X}}}) + \tilde{O}(\sqrt{Tm \gamma_{Tm}\log(\abs*{\mathcal{X}})}).
\end{align*}
\normalsize

\end{corollary}
\begin{proof}
    By Lemma \ref{lemma:appendix_initialization_bdd} in the appendix, we see that with a two-stage procedure where in the initialization stage we follow the maximal variance initialization strategy, we have $        \rho_m \leq \exp\left(\frac{m\gamma_{T_{init}}}{T_{init}}\right).$
    As noted in Lemma \ref{lemma:appendix_initialization_bdd}, whenever $T_{init}$ grows sublinearly with $T_{init}$, by picking $T_{init}$ sufficiently large, the term $\frac{m\gamma_{T_{init}}}{T_{init}}$ can be made to be $O(1)$, which means that $\rho_m$ can be bound by an absolute constant. We note that the first $T_{init}$ steps yield a regret of at most $\tilde{O}(T_{init} \log \abs*{\mathcal{X}})$, since the regret at each step can be bounded by $\xpt{\max_{x,x' \in \mathcal{X}} f(x) - f(x') }$, where without loss of generality, we may assume that for each $(x,x')$ pair, $f(x) - f(x') \sim N(0,2)$ (recall our initial assumption on $f \sim GP(0, k)$ where $\|k\|_\infty \leq 1$). Then, since the expectation of the maximum of N (possibly correlated) Gaussians each with variance bounded by 1 is at most $\sqrt{2\log N}$, it follows that $\xpt{\max_{x,x' \in \mathcal{X}} f(x) - f(x') } \leq \sqrt{2\log(\abs*{\mathcal{X}}^2)} = \sqrt{4\log \abs*{\mathcal{X}}}$. Thus, the first $T_{init}$ stages yields a regret of at most $\tilde{O}(T_{init} \sqrt{\abs*{\mathcal{X}}})$. Hence, by using the bound in Theorem \ref{theorem:main_result}, the overall regret across both phases thus satisfies 
\footnotesize
\begin{align*}
    \xpt{R_{T_{init}}} + \xpt{R_{T,m}} = \tilde{O}(T_{init}) + \tilde{O}(\sqrt{Tm \gamma_{Tm}\log(\abs*{\mathcal{X}})}).
\end{align*}
\normalsize
The specific choices of $T_{init}$ for the linear, exponential and Matern kernels follows from the bounds in their maximal information gain $\gamma_{\tau}$ term as specified in Lemma \ref{lemma:MIG_kernel_bounds}. For brevity, we do not go through the details in all three cases, focusing only on the squared exponential case. In this case, suppose without loss of generality that $\gamma_{\tau} = \log(\tau)^{d+1}$ exactly, with no constants in front of the polylogarithmic term. Then, it can be verified that by picking $T_{init} = m (\log(m^c))^{d+1}$ for a sufficiently large constant $c > 0$ depending on the dimension $d$, we have that 
\footnotesize
\begin{align}
    \frac{m \gamma_{T_{init}}}{T_{init}} = \frac{m (\log(T_{init}))^{d+1}}{T_{init}} = \frac{m (\log ( m  (\log(m^c))^{d+1}))^{d+1}}{m (\log(m^c))^{d+1}} =  \frac{(\log ( m  (\log(m^c))^{d+1}))^{d+1}}{(\log(m^c))^{d+1}}. \label{eq:sq_exp_T_init_detailed_derivation}
\end{align}
\normalsize
It can be verified that by picking $c$ sufficiently large, we have 
$m (\log(m^c))^{d+1} \leq m^c,$
in which case the term in (\ref{eq:sq_exp_T_init_detailed_derivation}) becomes less than 1. Similar analysis holds for the linear kernel, and for the Matern kernel, it can be verified that $T_{init}$ can be chosen to be on the order of $\operatorname{poly}(m)$. Finally, for the linear and squared exponential kernels, since $\gamma_{Tm} = O(\operatorname{polylog}(Tm))$ in these cases, the overall regret is $\tilde{O}(\sqrt{d Tm \log(\abs*{\mathcal{X}})})$ and $\tilde{O}(\sqrt{Tm (\log(Tm))^d \log(\abs*{\mathcal{X}})})$ respectively.
\end{proof}

\section{More details about experimental setup}
\label{appendix:simulation_details}

Our detailed experimental setup is as follows. For the GP prior (except the cases with known GP prior), we use the Matern kernel with $\nu$ parameter set as $\nu = 1.5$. For the likelihood noise, we set $\epsilon \sim N(0, \sigma_n^2)$, where $\sigma_n = 0.001$. We compute the performance of the algorithms across 10 runs, where for each run, each algorithm has access to the same random initialization dataset with 15 samples. Finally, we note that in a practical implementation of our algorithm, for any given $t$ and $i \in [m]$, it may happen that $\tilde{f}_{t,i}^* < \mu_t(x)$, in which case the algorithm will simply pick out the action $x$ with the highest $\mu_t(x)$. While such a situation does not affect the theoretical convergence, for better empirical performance that encourages more diversity, we resample $\tilde{f}_{t,i}^*$ whenever $\tilde{f}_{t,i}^* < \max_x \mu_t(x)$, until $\tilde{f}_{t,i}^* > \max_x \mu_t(x)$. The specific kernel, lengthscale and domain we used in the experiments for each of the test functions can be found in Tables \ref{tab:setup_2d_3d}, \ref{tab:setup_highD} and \ref{tab:setup_real} below.

\vspace{-3mm}
\begin{table}[htbp]
\centering
\tiny
\caption{Experimental set up for 2D/3D synthetic functions}
\label{tab:setup_2d_3d}
\begin{tabular}{|c@{\hskip 0pt}|c@{\hskip 0pt}|c@{\hskip 0pt}|c@{\hskip 0pt}|c@{\hskip 0pt}|c@{\hskip 0pt}|c@{\hskip 0pt}|}
\hline
 & Ackley-2D & Rosenbrock-2D & Bird-2D & Ackley-3d & GP-RBF-prior-2D & GP-RBF-prior-3D \\
\hline
Domain & $[-5,5]^2$ & $[-2,2] \times [-1,3]$ &$[-2\pi,2\pi]^2$  & $[-5,5]^3$ & $[-5,5]^2$ & $[0,1]^3$  \\
Lengthscale& $\ln(2)$ & $\ln(2)$ & $\ln(2)$ & $\ln(2) $ & 0.25 &  0.15 \\
Kernel & Matern & Matern  & Matern& Matern & RBF & RBF \\
Noise $\sigma$ & $10^{-3}$ &  $10^{-3}$ &  $10^{-3}$ &  $10^{-3}$  &  $10^{-3}$  &  $10^{-3}$  \\
\hline
\end{tabular}
\end{table}
\normalsize
\vspace{-4mm}

\begin{table}[h]
\centering
\tiny
\caption{Experimental set up for higher-dimensional functions}
\label{tab:setup_highD}
\begin{tabular}{|c@{\hskip 0pt}|c@{\hskip 0pt}|c@{\hskip 0pt}|c@{\hskip 0pt}|c@{\hskip 0pt}|c@{\hskip 0pt}|c@{\hskip 0pt}|}
\hline
 & Hartmann-6D & Griewank-8D & Michalewicz-10D  \\
\hline
Domain & $[0,1]^6$ & $[-1,4]^8$ &$[0,\pi]^{10}$   \\
Lengthscale  & $\ln(2)$ & $\ln(2)$ & $\ln(2)$  \\
Kernel & Matern & Matern & Matern \\
Noise $\sigma$ & $10^{-3}$ &  $10^{-3}$ &  $10^{-3}$   \\
\hline
\end{tabular}
\end{table}
\normalsize

\begin{table}[h]
\centering
\tiny
\caption{Experimental set up for real-world functions}
\label{tab:setup_real}
\begin{tabular}{|c@{\hskip 0pt}|c@{\hskip 0pt}|c@{\hskip 0pt}|c@{\hskip 0pt}|c@{\hskip 0pt}|c@{\hskip 0pt}|c@{\hskip 0pt}|}
\hline
 & Boston Housing (NN regression) & Robot-3D & Robot-4D \\
\hline
& $[1,100]\! \!\times\!\![0.001,0.1]\!\! \times\!\! [0.1,0.5]^2$ & $[-5,5]^2 \!\!\times\!\! [1,30]$ & $[-5,5]^2 \!\times\! [1,30]\!\! \times\! \![0,2\pi]$  \\
Lengthscale  &  $[0.1, 0.005, 0.1,0.1] $ & $\ln(2)$ &  $\ln(2)$ \\
Kernel  & Matern & Matern & Matern \\
Noise $\sigma$  &  $10^{-3}$  &  $10^{-3}$  &  $10^{-3}$  \\
\hline
\end{tabular}
\end{table}
\normalsize
\newpage
\bibliographystyle{siamplain}
\vspace{-4mm}
\bibliography{references.bib}

\begin{thebibliography}{10}

\bibitem{adachi_sober_2023}
{\sc M.~Adachi, S.~Hayakawa, S.~Hamid, M.~Jørgensen, H.~Oberhauser, and M.~A. Osborne}, {\em {SOBER}: {Highly} {Parallel} {Bayesian} {Optimization} and {Bayesian} {Quadrature} over {Discrete} and {Mixed} {Spaces}}, July 2023, \url{http://arxiv.org/abs/2301.11832} (accessed 2023-11-06).
\newblock arXiv:2301.11832 [cs, math, stat].

\bibitem{ament2024unexpected}
{\sc S.~Ament, S.~Daulton, D.~Eriksson, M.~Balandat, and E.~Bakshy}, {\em Unexpected improvements to expected improvement for bayesian optimization}, Advances in Neural Information Processing Systems, 36 (2024).

\bibitem{azimi_batch_nodate}
{\sc J.~Azimi, A.~Fern, and X.~Z. Fern}, {\em Batch {Bayesian} {Optimization} via {Simulation} {Matching}},  (2010).

\bibitem{baek2023ts}
{\sc J.~Baek and V.~Farias}, {\em Ts-ucb: Improving on thompson sampling with little to no additional computation}, in International Conference on Artificial Intelligence and Statistics, PMLR, 2023, pp.~11132--11148.

\bibitem{contal2013parallel}
{\sc E.~Contal, D.~Buffoni, A.~Robicquet, and N.~Vayatis}, {\em Parallel gaussian process optimization with upper confidence bound and pure exploration}, in Joint European Conference on Machine Learning and Knowledge Discovery in Databases, Springer, 2013, pp.~225--240.

\bibitem{dai2020federated}
{\sc Z.~Dai, B.~K.~H. Low, and P.~Jaillet}, {\em Federated bayesian optimization via thompson sampling}, Advances in Neural Information Processing Systems, 33 (2020), pp.~9687--9699.

\bibitem{daxberger2017distributed}
{\sc E.~A. Daxberger and B.~K.~H. Low}, {\em Distributed batch gaussian process optimization}, in International conference on machine learning, PMLR, 2017, pp.~951--960.

\bibitem{de_palma_sampling_2019}
{\sc A.~De~Palma, C.~Mendler-Dünner, T.~Parnell, A.~Anghel, and H.~Pozidis}, {\em Sampling {Acquisition} {Functions} for {Batch} {Bayesian} {Optimization}}, Oct. 2019, \url{http://arxiv.org/abs/1903.09434} (accessed 2023-11-12).
\newblock arXiv:1903.09434 [cs, stat].

\bibitem{desautels2014parallelizing}
{\sc T.~Desautels, A.~Krause, and J.~W. Burdick}, {\em Parallelizing exploration-exploitation tradeoffs in gaussian process bandit optimization}, Journal of Machine Learning Research, 15 (2014), pp.~3873--3923.

\bibitem{frazier2018tutorial}
{\sc P.~I. Frazier}, {\em A tutorial on bayesian optimization}, arXiv preprint arXiv:1807.02811,  (2018).

\bibitem{garcia2019fully}
{\sc J.~Garcia-Barcos and R.~Martinez-Cantin}, {\em Fully distributed bayesian optimization with stochastic policies}, arXiv preprint arXiv:1902.09992,  (2019).

\bibitem{garcia-barcos_fully_2019}
{\sc J.~Garcia-Barcos and R.~Martinez-Cantin}, {\em Fully {Distributed} {Bayesian} {Optimization} with {Stochastic} {Policies}}, July 2019, \url{http://arxiv.org/abs/1902.09992} (accessed 2024-01-07).
\newblock arXiv:1902.09992 [cs, stat].

\bibitem{garrido2019predictive}
{\sc E.~C. Garrido-Merch{\'a}n and D.~Hern{\'a}ndez-Lobato}, {\em Predictive entropy search for multi-objective bayesian optimization with constraints}, Neurocomputing, 361 (2019), pp.~50--68.

\bibitem{ginsbourger2008multi}
{\sc D.~Ginsbourger, R.~Le~Riche, and L.~Carraro}, {\em A multi-points criterion for deterministic parallel global optimization based on gaussian processes},  (2008).

\bibitem{gong2019quantile}
{\sc C.~Gong, J.~Peng, and Q.~Liu}, {\em Quantile stein variational gradient descent for batch bayesian optimization}, in International Conference on machine learning, PMLR, 2019, pp.~2347--2356.

\bibitem{gonzalez_batch_nodate}
{\sc J.~Gonzalez, Z.~Dai, P.~Hennig, and N.~Lawrence}, {\em Batch {Bayesian} {Optimization} via {Local} {Penalization}},  (2015).

\bibitem{hennig2012entropy}
{\sc P.~Hennig and C.~J. Schuler}, {\em Entropy search for information-efficient global optimization.}, Journal of Machine Learning Research, 13 (2012).

\bibitem{hernandez2015predictive}
{\sc J.~M. Hern{\'a}ndez-Lobato, M.~Gelbart, M.~Hoffman, R.~Adams, and Z.~Ghahramani}, {\em Predictive entropy search for bayesian optimization with unknown constraints}, in International conference on machine learning, PMLR, 2015, pp.~1699--1707.

\bibitem{hernandez2017parallel}
{\sc J.~M. Hern{\'a}ndez-Lobato, J.~Requeima, E.~O. Pyzer-Knapp, and A.~Aspuru-Guzik}, {\em Parallel and distributed thompson sampling for large-scale accelerated exploration of chemical space}, in International conference on machine learning, PMLR, 2017, pp.~1470--1479.

\bibitem{hunt2020batch}
{\sc N.~Hunt}, {\em Batch Bayesian optimization}, PhD thesis, Massachusetts Institute of Technology, 2020.

\bibitem{hvarfner2022joint}
{\sc C.~Hvarfner, F.~Hutter, and L.~Nardi}, {\em Joint entropy search for maximally-informed bayesian optimization}, Advances in Neural Information Processing Systems, 35 (2022), pp.~11494--11506.

\bibitem{kandasamy2018parallelised}
{\sc K.~Kandasamy, A.~Krishnamurthy, J.~Schneider, and B.~P{\'o}czos}, {\em Parallelised bayesian optimisation via thompson sampling}, in International Conference on Artificial Intelligence and Statistics, PMLR, 2018, pp.~133--142.

\bibitem{kaufmann2012bayesian}
{\sc E.~Kaufmann, O.~Capp{\'e}, and A.~Garivier}, {\em On bayesian upper confidence bounds for bandit problems}, in Artificial intelligence and statistics, PMLR, 2012, pp.~592--600.

\bibitem{kirschner2018information}
{\sc J.~Kirschner and A.~Krause}, {\em Information directed sampling and bandits with heteroscedastic noise}, in Conference On Learning Theory, PMLR, 2018, pp.~358--384.

\bibitem{letham2019constrained}
{\sc B.~Letham, B.~Karrer, G.~Ottoni, and E.~Bakshy}, {\em Constrained bayesian optimization with noisy experiments},  (2019).

\bibitem{ma2023gaussian}
{\sc H.~Ma, T.~Zhang, Y.~Wu, F.~P. Calmon, and N.~Li}, {\em Gaussian max-value entropy search for multi-agent bayesian optimization}, arXiv preprint arXiv:2303.05694,  (2023).

\bibitem{nava_diversified_2022}
{\sc E.~Nava, M.~Mutný, and A.~Krause}, {\em Diversified {Sampling} for {Batched} {Bayesian} {Optimization} with {Determinantal} {Point} {Processes}}, Feb. 2022, \url{http://arxiv.org/abs/2110.11665} (accessed 2024-02-19).
\newblock arXiv:2110.11665 [cs, stat].

\bibitem{russo2014learning}
{\sc D.~Russo and B.~Van~Roy}, {\em Learning to optimize via information-directed sampling}, Advances in Neural Information Processing Systems, 27 (2014).

\bibitem{shah2015parallel}
{\sc A.~Shah and Z.~Ghahramani}, {\em Parallel predictive entropy search for batch global optimization of expensive objective functions}, Advances in neural information processing systems, 28 (2015).

\bibitem{srinivas2009gaussian}
{\sc N.~Srinivas, A.~Krause, S.~M. Kakade, and M.~Seeger}, {\em Gaussian process optimization in the bandit setting: No regret and experimental design}, arXiv preprint arXiv:0912.3995,  (2009).

\bibitem{takeno2020multi}
{\sc S.~Takeno, H.~Fukuoka, Y.~Tsukada, T.~Koyama, M.~Shiga, I.~Takeuchi, and M.~Karasuyama}, {\em Multi-fidelity bayesian optimization with max-value entropy search and its parallelization}, in International Conference on Machine Learning, PMLR, 2020, pp.~9334--9345.

\bibitem{verma2022bayesian}
{\sc A.~Verma, Z.~Dai, and B.~K.~H. Low}, {\em Bayesian optimization under stochastic delayed feedback}, in International Conference on Machine Learning, PMLR, 2022, pp.~22145--22167.

\bibitem{wang2017max}
{\sc Z.~Wang and S.~Jegelka}, {\em Max-value entropy search for efficient bayesian optimization}, in International Conference on Machine Learning, PMLR, 2017, pp.~3627--3635.

\bibitem{wang2016optimization}
{\sc Z.~Wang, B.~Zhou, and S.~Jegelka}, {\em Optimization as estimation with gaussian processes in bandit settings}, in Artificial Intelligence and Statistics, PMLR, 2016, pp.~1022--1031.

\bibitem{williams2006gaussian}
{\sc C.~K. Williams and C.~E. Rasmussen}, {\em Gaussian processes for machine learning}, vol.~2, MIT press Cambridge, MA, 2006.

\bibitem{zhan2020expected}
{\sc D.~Zhan and H.~Xing}, {\em Expected improvement for expensive optimization: a review}, Journal of Global Optimization, 78 (2020), pp.~507--544.

\end{thebibliography}

\end{document}